\documentclass[10pt,twocolumn,letterpaper]{article}

\usepackage{cvpr}              %

\usepackage[dvipsnames]{xcolor}

\newcommand{\equallyspacedcaption}[4]{
  \hspace*{#3}
  #1
  \hspace*{\fill}
  \foreach \value in {#2} {
    \value
    \hspace*{\fill} %
  }
  \hspace*{#4}
}

\newcommand{\anycolfig}[8]{
 \begin{figure*}
    \hspace*{-0.5cm} %
    \begin{tabular}{*{#1}{c}}
      \foreach \image/\name in {#3} {
        \begin{subfigure}{\dimexpr\textwidth/#1\relax}
          \centering
          \caption*{\equallyspacedcaption{\name}{#4}{#7}{#8}} %
          \includegraphics[width=1.0\linewidth]{#2/\image} %
        \end{subfigure}
      }
    \end{tabular}
    \vspace{-0.3cm}
    \caption{#6}
    \label{#5}
    \vspace{-0.3cm}
  \end{figure*}
}

\newcommand{\anycolfigpdf}[8]{
 \begin{figure*}
    \hspace*{-0.5cm} %
    \begin{tabular}{*{#1}{c}}
      \foreach \image/\page/\name in {#3} {
        \begin{subfigure}{\dimexpr\textwidth/#1\relax}
          \centering
          \caption*{\equallyspacedcaption{\name}{#4}{#7}{#8}} %
          \includegraphics[page=\page,width=1.0\linewidth]{#2/\image} %
        \end{subfigure}
      }
    \end{tabular}
    \vspace{-0.3cm}
    \caption{#6}
    \label{#5}
    \vspace{-0.3cm}
  \end{figure*}
}

\usepackage{url}            %
\usepackage{booktabs}       %
\usepackage{amsfonts}       %
\usepackage{nicefrac}       %
\usepackage{microtype}      %
\usepackage{xcolor}         %
\usepackage{pgfplots}
\usepackage{amsmath}
\usepackage[numbers]{natbib}
\usepackage{graphicx,subcaption,mwe}

\newcommand{\guidancename}{Readout Guidance}
\newcommand{\guidancenamelower}{readout guidance}
\newcommand{\methodname}{Readout Guidance}

\newcommand{\sdvonefive}{SDv1-5}
\newcommand{\sdxl}{SDXL}

\newcommand{\promptcaption}[1]{\textit{``#1''}}

\usepackage{pgfplots}
\usepackage{graphicx}
\usepackage{subcaption}

\definecolor{cvprblue}{rgb}{0.21,0.49,0.74}
\usepackage[pagebackref,breaklinks,colorlinks,citecolor=cvprblue]{hyperref}
\hypersetup{linkcolor=black} %

\def\paperID{8617} %
\def\confName{CVPR}
\def\confYear{2024}

\title{Readout Guidance: Learning Control from Diffusion Features}

\author{Grace Luo$^{1, 2}$
\and
\hspace{-1em} Trevor Darrell$^{2}$
\and
\hspace{-1em} Oliver Wang$^{1}$
\and
\hspace{-1em} Dan B Goldman$^{1}$
\and
\hspace{-1em} Aleksander Holynski$^{1, 2}$ \\
\vspace{-1em}
\and
$^{1}$Google Research
\and
$^{2}$UC Berkeley
}

\begin{document}

\twocolumn[{%
\renewcommand\twocolumn[1][]{#1}%
\maketitle
\begin{center}
    \centering
    \vspace{-0.5cm}
    \captionsetup{type=figure}
    \includegraphics[width=\textwidth]{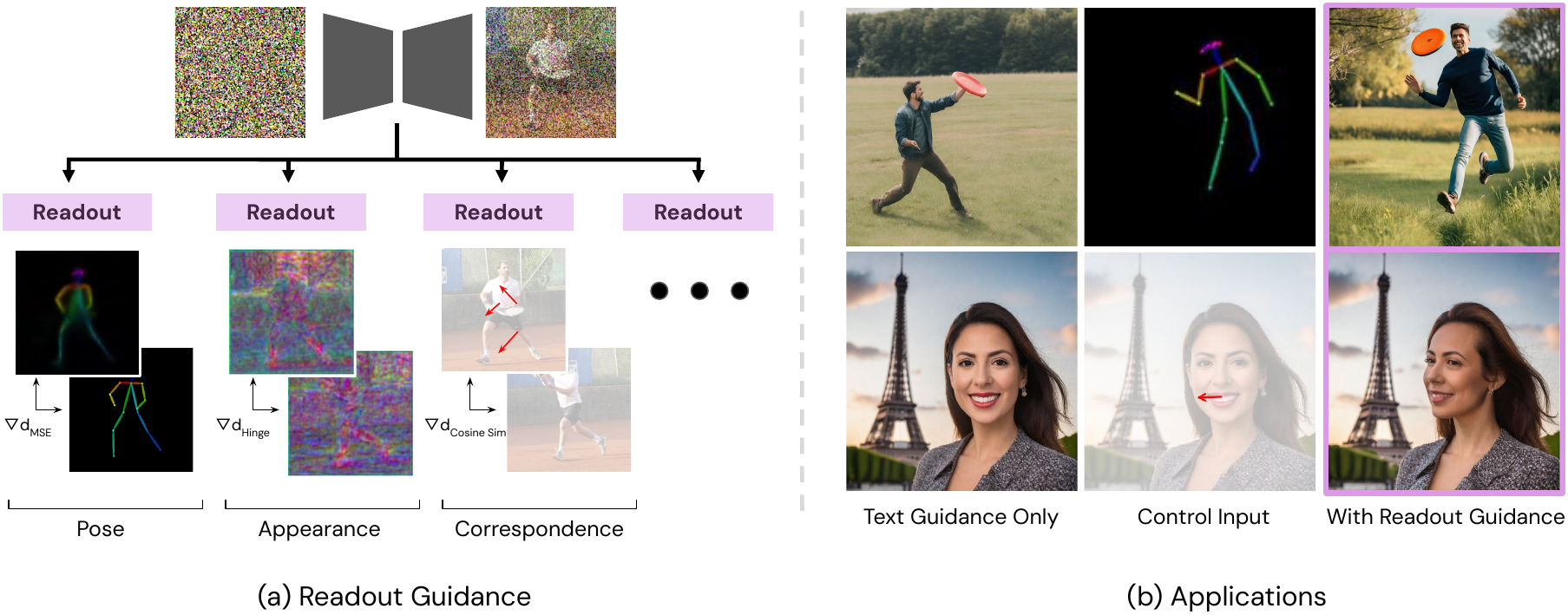}
    \captionof{figure}{\label{fig:teaser}Given a frozen pre-trained text-to-image diffusion model~\cite{Rombach_2022_CVPR}, we learn parameter-efficient \textit{readout heads} to interpret relevant signals, or \textit{readouts}, from the intermediate network features. These readouts can be single-image concepts such as pose and depth, or relative concepts between two images, such as appearance similarity and correspondence. We use the readouts for sampling-time guidance to enable controlled image generation.}
\end{center}%
}]

\maketitle

\begin{abstract}
\vspace{-1cm}
We present Readout Guidance, a method for controlling text-to-image diffusion models with learned signals. Readout Guidance uses \emph{readout heads}, lightweight networks trained to extract signals from the features of a pre-trained, frozen diffusion model at every timestep. These readouts can encode single-image properties, such as pose, depth, and edges; or higher-order properties that relate multiple images, such as correspondence and appearance similarity. Furthermore, by comparing the readout estimates to a user-defined target, and back-propagating the gradient through the readout head, these estimates can be used to guide the sampling process. Compared to prior methods for conditional generation, Readout Guidance requires significantly fewer added parameters and training samples, and offers a convenient and simple recipe for reproducing different forms of conditional control under a single framework, with a single architecture and sampling procedure. We showcase these benefits in the applications of drag-based manipulation, identity-consistent generation, and spatially aligned control.
Project page: \small{\emph{\url{https://readout-guidance.github.io}}}.
\end{abstract}

\section{Introduction}\label{sec:intro}
Diffusion models have shown enormous potential in accurately modeling the space of natural images. However, one remaining open challenge that is critical to many applications is enabling arbitrary user control over their outputs.
Existing solutions for enabling custom user control typically involve substantial model training on large annotated datasets, a process that is cumbersome and often infeasible for the average user.
In this work, we provide an alternative solution for enabling user control that combines two ideas; first, that diffusion models contain rich internal representations that are useful for extracting relevant image properties, and second, that these extracted image properties can be used to guide the generation process towards desired user constraints. We call the combined approach~\guidancename, because it makes use of small auxiliary readout heads that can be easily trained on top of a frozen diffusion model.

Taking as input the set of intermediate diffusion features, these readout heads can be trained to extract arbitrary properties about the image being generated (\autoref{fig:teaser}, left). These can include image-space properties, such as human pose, depth maps, and edges; but can also be higher-order properties that relate two or more images, such as appearance similarity, correspondence, or shared identity. These readout heads consist of very few parameters, meaning that they can be trained on a single consumer GPU in a matter of hours---and since they bootstrap the already-rich diffusion features, they only need as few as 100 training examples. 
Beyond their utility in efficiently extracting properties about the generated images, the readout heads offer a useful mechanism for \emph{guiding} the sampling process (\autoref{fig:teaser}, right). At each step of the sampling process, properties can be extracted from the readout heads and compared to user-defined targets. In a similar fashion to classifier guidance~\cite{dhariwal2021diffusion}, this comparison can be used as a guidance signal that encourages the generated image to match the target constraints. %

We show that \guidancename{} can implement a number of popular forms of user control---all within the same simple guidance framework. 
In particular, we demonstrate state-of-the-art performance on the task of drag-based image manipulation, an application that has previously required bespoke architectural modifications and additional per-example fine-tuning.
We also showcase our method on the task of identity-consistent image generation, in which outputs can be guided to contain the same person as a reference image.
Finally, our method can also be used for spatially-aligned controls, such as depth-guided or pose-guided generation, as popularized by ControlNet~\cite{zhang2023adding} and T2IAdapter~\cite{mou2023t2i}. 
Notably, when compared to ControlNet~\cite{zhang2023adding}, our method requires significantly less training data (as few as 100 supervised pairs vs. 200k), much less training time (a few hours vs. more than a week), and fewer added parameters (49MB vs. 1.4GB)~\cite{controlpose}.

\anycolfig{2}{figures/drag}{
  generated_1.pdf/,
  generated_2.pdf/
}{}{fig:dragdiff_synthetic}{\textbf{Drag Based Manipulation (Generated Images)}: We show generated images with a single user correspondence constraint (with overlay) followed by the~\guidancename{} generated result. Please see the Supplemental for the associated text prompts.}{0cm}{0cm}

\section{Related Work}\label{sec:related}
\textbf{Conditional Diffusion Models.} It has become increasingly popular to fine-tune text-to-image diffusion models to condition generation on signals beyond text, including camera pose~\cite{liu2023zero}, a reference identity~\cite{ruiz2023dreambooth}, a reference image~\cite{sdinpaint, brooks2023instructpix2pix, sdimagevar}, a depth map~\cite{sddepth}, and more~\cite{lhhuang2023composer}. 
Due to the high cost of training diffusion models, many methods propose to keep the base model frozen and instead train an additional network that takes in the control signal and modulates the intermediate diffusion features accordingly. 
These types of models, including ControlNets~\cite{zhang2023adding}, Adapters~\cite{mou2023t2i}, and LoRAs~\cite{hu2021lora}, require less compute to train, fewer training samples, and can be combined in various ways because they build on the same frozen base model. 
Our method is orthogonal and complementary to this kind of adapter tuning, since our method aims to~\textit{guide} the sampling process based on the diffusion model's features, rather than modulate it, and can therefore be applied to any base model with any additional adapter.
Our approach is more similar to classifier guidance, which is shown in Dhariwal \& Nichol~\cite{dhariwal2021diffusion} to not only enable conditional generation from unconditional models, but also reinforce the capabilities of a conditional model. In our experiments, we similarly demonstrate that our method can be applied to these adapter-based models~\cite{zhang2023adding, mou2023t2i} to further improve their control capabilities.\\
\textbf{Sampling-Time Guidance.} Because diffusion models synthesize images through an iterative sampling process rather than a single forward pass, one can guide the sampling process in a particular direction without modifying the base model.
Such guidance was first achieved with gradients from an ImageNet~\cite{deng2009imagenet} classifier trained on noisy images from the diffusion forward process~\cite{dhariwal2021diffusion}. 
Follow-up works explored alternative guidance approaches including the difference of score estimates between a conditional and unconditional model~\cite{ho2021classifierfree}, off-the-shelf models that operate on the predicted clean image~\cite{bansal2023universal, wallace2023endtoend}, and hand-designed training-free functions that operate on diffusion features~\cite{park2024shape, chen2023training, lian2023llmgrounded, epstein2023selfguidance}. In our work, we focus on \textit{learning} this guidance function from diffusion features. Depth-Aware Guidance~\cite{kim2022dag} first explored a similar idea for depth refinement, Sketch-Guided Diffusion~\cite{voynov2022sketch} for edge guidance, and Mid-U Guidance~\cite{whitaker2023midu} for aesthetic guidance. We expand this line of work into a more general form that can be used for non spatially-aligned applications such as drag-based manipulation.
\\
\textbf{Diffusion Model Representations.} 
There exist works that demonstrate the rich and expressive nature of diffusion features beyond image synthesis, and how they can be readily applied to tasks like segmentation~\cite{baranchuk2021label, xu2022odise, zhao2023unleashing, wu2023datasetdm}, depth estimation~\cite{zhao2023unleashing, wu2023datasetdm}, human pose prediction~\cite{wu2023datasetdm}, and semantic correspondence~\cite{luo2023dhf, hedlin2023unsupervised, zhang2023tale, tang2023dift}. These methods typically train additional decoders to extract an application-specific representation from the features of a particular, single diffusion timestep. 
To use these signals for guidance, we build on this work to instead extract an evolving prediction of these image properties at \emph{each} step of the \textit{entire} diffusion process.
\\
\textbf{Drag-based Manipulation.} The concurrent works DragDiffusion~\cite{shi2023dragdiffusion} and DragonDiffusion~\cite{mou2023dragondiffusion} explore drag-based editing~\cite{pan2023drag} with diffusion models. 
In this task, the user provides an image and a sparse set of point correspondences and the method produces an edited image that respects the specified appearance and deformation. These works enforce their appearance constraint using custom sampling-time processes which include single-image finetuning~\cite{shi2023dragdiffusion} or shared attention features~\cite{mou2023dragondiffusion} (following work in video stylization~\cite{wu2023tune, khachatryan2023text2video, cao2023masactrl, geyer2023tokenflow}). The deformation constraints are enforced by encouraging high similarity between the corresponding points in the feature maps extracted from the diffusion model, as proposed in diffusion-based semantic correspondence methods~\cite{tang2023dift,luo2023dhf}. 
In contrast, our method learns the appearance similarity and correspondence constraints automatically from video data and applies both through the same sampling-time guidance procedure without any hand-designed operators or per-example fine-tuning.

\section{Readout Guidance}\label{sec:guidance}
In this section, we describe the process of Readout Guidance: first describing the use of readout heads in the sampling process, then providing a general recipe for training a readout head on a custom dataset. In Sec.~\ref{sec:heads}, we describe a number of example readout heads, showing that they can model spatially aligned properties like depth and human pose, as well as pairwise relative properties, such as correspondence and appearance similarity. We later showcase these readout heads in a number of conditional control applications, such as drag-based manipulation (\autoref{fig:dragdiff_synthetic},~\ref{fig:dragdiff_real}), appearance preservation~(\autoref{fig:image_var},~\ref{fig:identity}), and spatially aligned control (\autoref{fig:controlnet}).
\begin{figure*}
    \centering
    \includegraphics[width=1.0\linewidth]{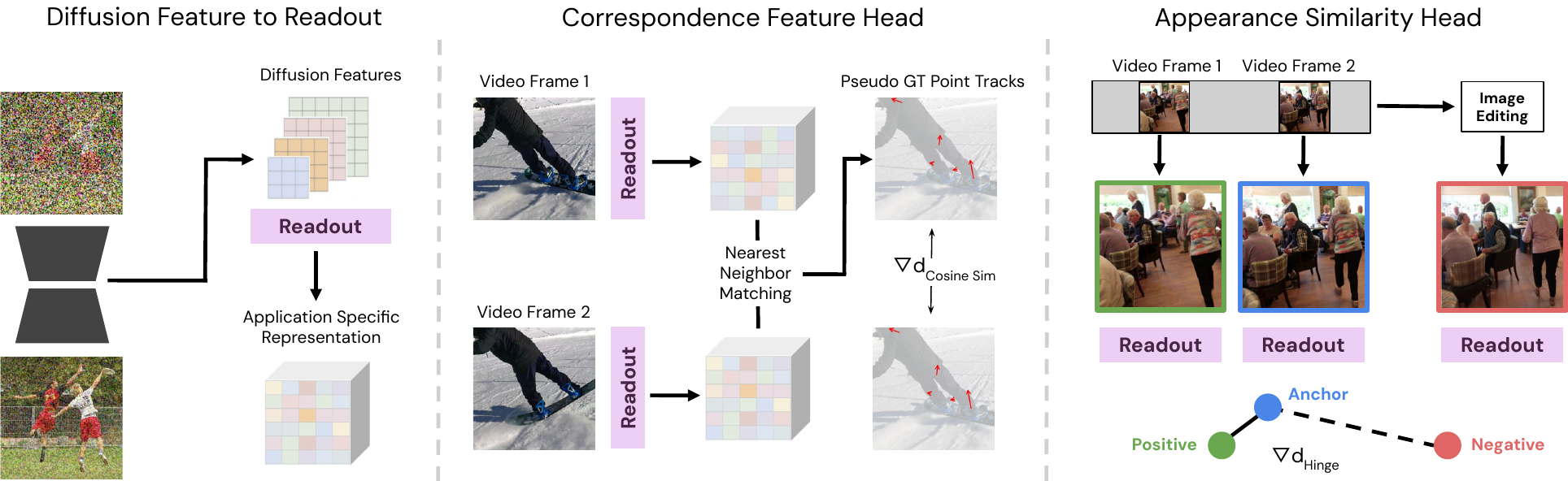}
    \caption{{\textbf{Readout Head Training}}: (left) Readout heads convert frozen diffusion features into representations useful for a diverse set of tasks,
    including predicting (middle) correspondences between a source and target image and (right) an appearance similarity feature between an anchor and positive / negative images.}
    \label{fig:approach_relative}
    \vspace{-0.3cm}
\end{figure*}
\subsection{Background}
Classifier guidance is a sampling procedure that enables conditional synthesis in an unconditional diffusion model.
For the deterministic DDIM sampler~\cite{song2020denoising}, Dhariwal \& Nichol~\cite{dhariwal2021diffusion} derive the update rule:
$$\hat{\epsilon}_t \gets \epsilon_{\theta}(x_t) - \sqrt{1 - \bar{\alpha}_t} \nabla_{x_t} \mathrm{log}\, p_{\phi}(y|x_t)$$
where $\epsilon_\theta(\cdot)$ is the diffusion model, $p_{\phi}(y|\cdot)$ is the classifier, and $\sqrt{1 - \bar{\alpha}_t}$ is a timestep-conditional scaling factor. 
One can think of this process as the classifier ``nudging'' the generation at each step such that it remains on the natural image manifold but moves towards to the classifier's criteria.
In our method, we bootstrap pre-trained diffusion features from the frozen U-Net decoder to train a small readout head on top of the diffusion model.
This allows us to design simple heads that are the same across entire classes of tasks (i.e., we use one architecture for all relative properties and another for all spatially-aligned properties) and only differ in the input data and loss.
\subsection{A Revised Guidance Recipe}
\textbf{From Classification to Regression.} In class and text guidance, the output domain $y$ of the guidance function is often a fixed set of classes (e.g., object categories or a binary decision of ``does the caption match'' vs. ``does the caption not match''). We replace classifiers with regressors that predict continuous values, where instead of maximizing a log probability, a distance function is minimized. As such, we write our guidance function as a distance between the reference $r$ and our predicted readout $\hat{r} = f_{\psi}(x_t)$.
$$\hat{\epsilon}_t \gets \epsilon_{\theta}(x_t) + w \cdot \nabla_{x_t} d(r, f_{\psi}(x_t))$$
where $\epsilon_{\theta}(\cdot)$ is the diffusion model, $w$ is the guidance weight, and $f_{\psi}(\cdot)$ is our learned readout head. \\
\textbf{Relative Constraints.} In the case of readout heads that relate multiple images, the reference is not a user input but rather derived from a separate diffusion process with a reference noisy image $z_t$: 
$$\hat{\epsilon}_t \gets \epsilon_{\theta}(x_t) + w \cdot \nabla_{x_t} d(f_{\psi}(z_t), f_{\psi}(x_t))$$
Our distance function $d(\cdot, \cdot)$ does not necessarily have to be computed between spatially aligned positions on the readouts -- it can encode arbitrary spatial relationships, such as corresponding points or an object to be re-positioned. This flexibility in the loss formulation represents the generality of our method over ControlNet~\cite{zhang2023adding}, which modulates features via a \textit{per-pixel sum} between the control and diffusion model features, thereby making the method difficult to adapt to spatially unaligned constraints.
\\
\noindent\textbf{Combining Guidance Functions.} One can also combine our~\guidancenamelower{} with other common forms of guidance, such as classifier-free text guidance~\cite{ho2021classifierfree}. It is also possible to guide with multiple readout heads $\psi_1, ..., \psi_n$ simultaneously. For example, we guide with both an appearance similarity and correspondence feature head for drag-based manipulation (\autoref{fig:dragdiff_synthetic}).

\anycolfig{3}{figures/controlnet}{
  pose.pdf/Pose Control,
  depth.pdf/Depth Control,
  edge.pdf/Edge Control
}{RG (Ours), Readout}{fig:controlnet}{\textbf{Spatially Aligned Controls (Generated Images)}: In each example, we show the input user control, provided as a pose, depth map, or edge map derived from a different image (not shown), as well as our generated result, and a visualization of the readout head output for the generated image. We show more results in the Supplemental.}{0.2cm}{0cm}

\section{Readout Heads}\label{sec:heads}
\subsection{Common Architecture}
Our readout heads extend the architecture of Diffusion Hyperfeatures~\cite{luo2023dhf}. The heads extract features from the frozen decoder layers of the U-Net, reshape them with learned projection layers and bilinear upsampling, and then learn mixing weights to compute a weighted average of the layer features.
Unlike fine-tuned conditional models~\cite{zhang2023adding, mou2023t2i}, our method does not require a large captioned dataset for training. We find that our method works well with extracted features from the unconditional branch, using the empty string ``'' as the prompt for all images.
Given that we are also using these heads for guidance at sampling time, unlike the multi-timestep aggregation in Diffusion Hyperfeatures, we instead aggregate only over network layers. This allows us to query readout estimates at each sampling timestep, so that they can be used for guidance. Thus, we add timestep conditioning to the readout heads, and train them on features resulting from noisy images seen during the diffusion forward process. 
We define two types of heads, one for spatially-aligned properties, such as depth and pose, and one for ``relative'' properties, where the goal is to enforce semantics of one frame \emph{relative} to another reference frame.
For spatially-aligned properties, we add three convolutional layers to convert the feature map into an RGB readout, and for relative properties like correspondence and appearance similarity, we compute a distance metric directly on the feature map produced by the readout head. \autoref{fig:approach_relative} depicts the training procedure of our relative heads, and a more detailed architecture diagram can be found in the Supplemental.
\subsection{Spatially Aligned Heads}\label{subsec:spatial}
\textbf{Pose, Edge, Depth Head.} We train our spatial heads on images from PascalVOC~\cite{pascal-voc-2012}. 
We use a similar setup to ControlNet~\cite{zhang2023adding}, in which pseudo labels for the control are computed by off-the-shelf models and standardized into RGB images; we use OpenPose~\cite{8765346} to extract human pose, MiDaS~\cite{Ranftl2021, Ranftl2022} for depth, and HED~\cite{xie2015holistically} for edges.
For a given image, the spatial head extracts an aligned readout image.
We then compute a per-pixel mean squared error loss against the ground truth pseudo label. 
For the pose and edge heads we re-weight the loss such that the contribution of the non-null and null regions are equal, as both controls are inherently sparse and include large portions of black background. In~\autoref{fig:controlnet} we depict example generations guided to match a user control by these heads as well as their associated readouts.
\subsection{Relative Heads}\label{subsec:relative}
\textbf{Correspondence Feature Head.} The goal of the correspondence feature head is to learn features that can be used to enforce correspondence constraints to the generated image.
We train this readout head on video frames on DAVIS, following the approach of Diffusion Hyperfeatures~\cite{luo2023dhf}, using image pairs with labeled point correspondences and training a network such that the feature distance between corresponding points is minimized, \textit{i.e.}, the target point feature is the nearest neighbor for a given source point feature. 
We compute pseudo-labels using a point tracking algorithm~\cite{karaev2023cotracker} to track a grid of query points across the entire video. 
We randomly select two frames from the same video and a subset of the tracked points that are visible in both frames. For each image in the pair, the correspondence feature head constructs an associated feature map. 
We then use a contrastive loss -- symmetric cross entropy~\cite{clip} -- to maximize the cosine similarity of matched source and target points. \\
\noindent\textbf{Appearance Similarity Head.} The goal of the appearance similarity head is to encourage the generation towards one where the appearance is similar to a reference image, but the image composition is unconstrained. 
That is, we want to penalize changes related to color, texture, and identity, but not changes related to object pose or camera angle. 
To do this, we define a loss based on \emph{videos}, where we pick an anchor frame randomly selected from a video, and use a triplet loss to enforce that its features are more similar to those from another random frame from the \emph{same video} than they are to an image with the \emph{same structural layout} as the anchor, but with a \emph{different} appearance. 
We achieve this by using an image editing method (SDEdit~\cite{meng2022sdedit, sdimg2img}) that noises and denoises the anchor frame to produce variations that are similar in structure and semantic content, but have variably perturbed textural content. 
All three images (anchor, positive, negative) are processed independently by the appearance head, which constructs corresponding feature maps.
For each pair of feature maps, we compute a per-pixel cosine distance and spatially pool these distances into a single scalar. We supervise the anchor to be closer in distance to the positive using a hinge loss with a margin of 0.5~\cite{gal2010largescale, fu2023learning}.
We train entirely on videos from DAVIS~\cite{DBLP:journals/corr/Pont-TusetPCASG17}.
\\
\noindent\textbf{Identity Head.} 
We also train a variant of the appearance similarity head with an identical architecture and training scheme but specific to preserving the identity of people. We train on tightly cropped images of faces from the CelebA-HQ dataset~\cite{karras2017progressive}, where we use the same image editing technique~\cite{meng2022sdedit} to produce hard negatives containing similar faces of different identities.

\anycolfig{2}{figures/drag}{
  real_1.pdf/Drag Control,
  real_2.pdf/Drag Control
}{DragDiffusion, RG (Ours)}{fig:dragdiff_real}{\textbf{Drag Based Manipulation (Real Images)}: The appearance similarity and correspondence feature head can operate on real images when seeding the reference features with those from DDIM inversion~\cite{song2020denoising}. We compare against the concurrent work DragDiffusion~\cite{pan2023drag}. Note that DragDiffusion requires an additional user input mask whereas our method does not.}{0.7cm}{-0.2cm}
\section{Results}\label{sec:results}
\begin{figure}
    \caption*{\equallyspacedcaption{Pose Control}{ControlNet, ControlNet + RG, Readout}{0cm}{0cm}}
    \vspace{-0.3cm}
    \includegraphics[width=1.0\linewidth]{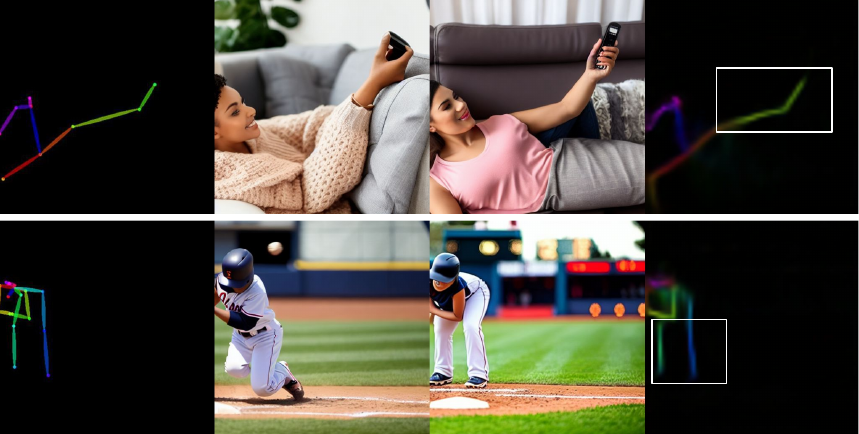} 
    \caption*{\equallyspacedcaption{Pose Control}{T2IAdapter, T2IAdapter + RG, Readout}{0cm}{0cm}}
    \vspace{-0.3cm}
    \includegraphics[width=1.0\linewidth]{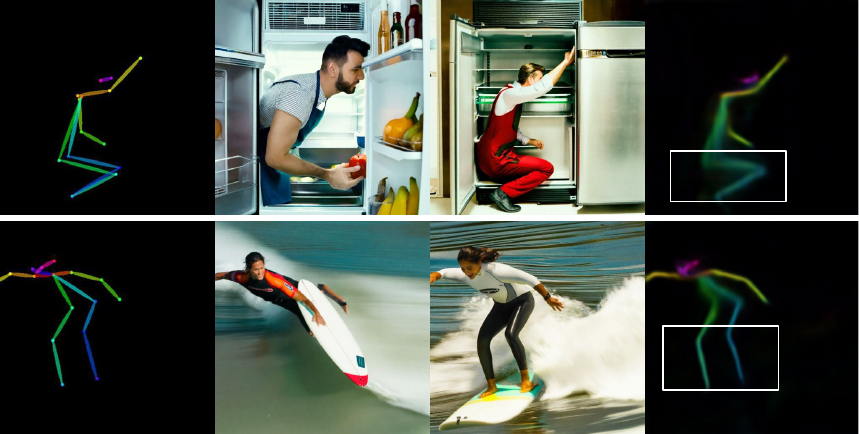}
    \vspace{-0.1cm}
    \caption{\textbf{Control Refinement}: Using Readout Guidance in combination with ControlNet~\cite{zhang2023adding} (top) and T2IAdapter~\cite{mou2023t2i} (bottom) can correct mistakes for difficult pose-guided generation cases shown here. We highlight these areas with overlaid white boxes.}
    \label{fig:controlnet_ours}
    \vspace{-0.6cm}
\end{figure}

\begin{figure}
    \caption*{\equallyspacedcaption{Pose Control}{100 Images, 1k Images, 8.5k Images}{0.1cm}{0cm}}
    \vspace{-0.3cm}
    \includegraphics[width=1.0\linewidth]{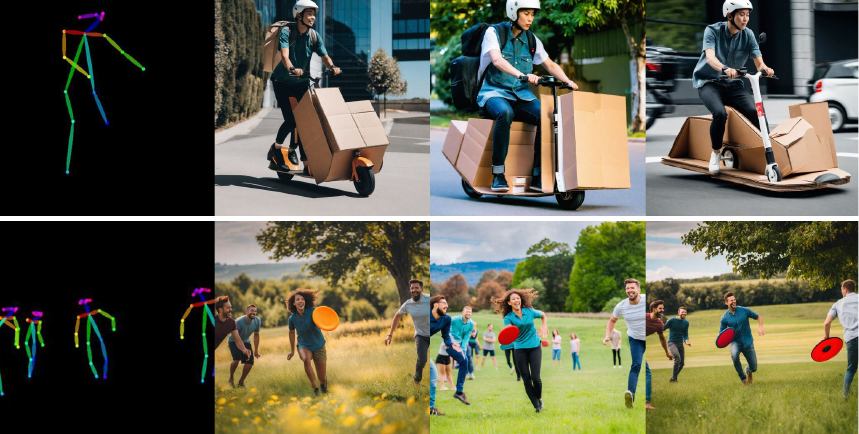} 
    \vspace{-0.1cm}
    \caption{\textbf{Limited Data}: Readout heads are lightweight enough to apply in domains where there is little supervised data to train from. As few as 100 examples are needed to train a pose control model.}
    \label{fig:train_size}
    \vspace{-0.5cm}
\end{figure}

\begin{figure}
\begin{tikzpicture}
  \begin{axis}[
    title={Training Data vs. Pose Control Performance (SDXL)},
    xlabel={Amount of Training Data},
    ylabel={PCK ($\alpha=0.05$)},
    y label style={at={(axis description cs:0.1,0.5)}, anchor=south},
    grid=both,
    xmode=log, %
    xtick={100,1000,8500,3000000}, %
    xticklabels={100,1k,8.5k,3M}, %
    ytick={10, 20, 30, 40, 50, 60},
    scaled x ticks=false, %
    grid style={gray, thin, dotted},
    width=8cm, %
    height=6cm, %
    ymin=0,
    ymax=65,
  ]
    \addplot[only marks, mark=diamond*, mark options={scale=1.5, fill=green, draw=none},
           nodes near coords, %
           point meta=explicit symbolic,
           visualization depends on={value \thisrow{label} \as \mylabel},
           nodes near coords style={yshift=-33pt, xshift=-33pt}, %
           ]
    table[x=x, y=y, meta=label] {
      x       y       label
      3008500 54.54   {\parbox{3cm}{\centering T2IAdapter + RG \\ (85M params)}}
    };
  \addplot[only marks, mark=square*, mark options={scale=1.5, fill=blue, draw=none},
           nodes near coords, %
           point meta=explicit symbolic,
           visualization depends on={value \thisrow{label} \as \mylabel},
           nodes near coords style={yshift=-34pt, xshift=3pt}, %
           ]
    table[x=x, y=y, meta=label] {
      x       y       label
      100     15.24   {}
      1000    19.65   {\parbox{3cm}{\centering RG \\ (5.9M params)}}
      8500    24.29   {}
    };
  \addplot[only marks, mark=triangle*, mark options={scale=1.5, fill=red, draw=none},
           nodes near coords, %
           point meta=explicit symbolic,
           visualization depends on={value \thisrow{label} \as \mylabel},
           nodes near coords style={yshift=-30pt, xshift=-25pt}, %
           ] 
    table[x=x, y=y, meta=label] {
      x       y       label
      3000000 24.06   {\parbox{3cm}{\centering T2IAdapter \\ (79M params)}}
    };
  \addplot[blue, thick] coordinates {(100, 15.24) (1000, 19.65) (8500, 24.29)};
  \end{axis}
\end{tikzpicture}
\vspace{-0.1cm}
\caption{\textbf{Readout Guidance performs comparably to a conditional model~\cite{t2iadapterpose} with 350x less training data}: We compute the percentage of correct keypoints (PCK) between an input pose and the pose of the synthesized image on 100 random images with humans from MSCOCO~\cite{lin2014microsoft}. 
See additional results in the Supplemental.}
\vspace{-0.5cm}
\label{fig:pose_plot}
\end{figure}
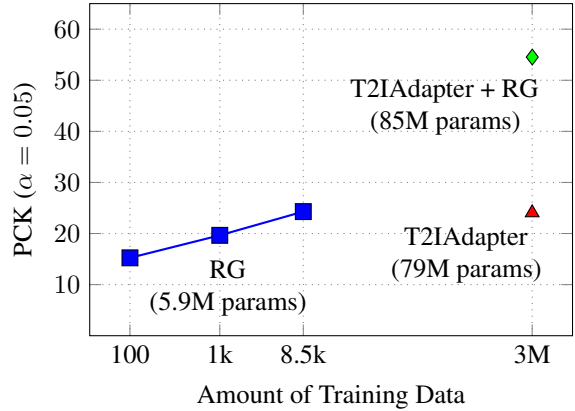

\textbf{Experimental Details.} We implement readout heads for both Stable Diffusion v1-5~\cite{Rombach_2022_CVPR} (\sdvonefive) and Stable Diffusion XL~\cite{podell2023sdxl} (\sdxl), such that we use the same model as the baseline for every comparison. 
We sample with image resolutions of 512 and 1024, resulting in latent resolutions of 64 and 128 respectively. Although these are both latent diffusion models~\cite{Rombach_2022_CVPR},
we find that our method produces high quality readouts and guidance operating directly on latent U-Net features, without the need for a latent decoder. In all our experiments, both when training our heads and sampling, we use a single Nvidia A100 40G GPU, and training a readout head takes at most three hours. We list additional information about the compute requirements and sampling hyperparameters for our method in the Supplemental.\\
\textbf{Drag-Based Manipulation.} 
Given a reference image, as well as a sparse set of points representing a desired deformation, the model is tasked with producing an edited image that respects both the input image appearance and the desired transformation. In~\autoref{fig:dragdiff_real} we compare against the concurrent work DragDiffusion~\cite{shi2023dragdiffusion}, which finetunes a LoRA~\cite{hu2021lora} for each reference image and at sampling time optimizes such that the correspondences between the raw feature maps~\cite{tang2023dift} match the points.
For our method, we guide with our appearance similarity and correspondence feature heads across the entire diffusion process. This enables large out-of-plane motions, as seen in the first row where our method is able to fully rotate the tiger's head, whereas DragDiffusion slightly translates its nose upwards. 
These types of 3D transformations tend to be particularly challenging for prior work, likely because raw features encode some amount of orientation and perspective~(e.g., if the nose is forward vs. side facing).
We find that our correspondence feature head produces more orientation-agnostic features, likely due to our video training data, which allows us to correctly deform shapes.
In addition, our appearance similarity guidance helps to keep the background consistent while editing the foreground object, which means that as opposed to DragDiffusion, our method does not require an additional input foreground mask.
\begin{figure*}
    \caption*{Weight = 0.0 \hspace{0.7cm} $\xrightarrow{\hspace{8.4cm}}$ \hspace{0.3cm} Weight = 1.0 \hspace{1.0cm} Reference}
    \vspace{-0.2cm}
    \includegraphics[width=1.0\linewidth]{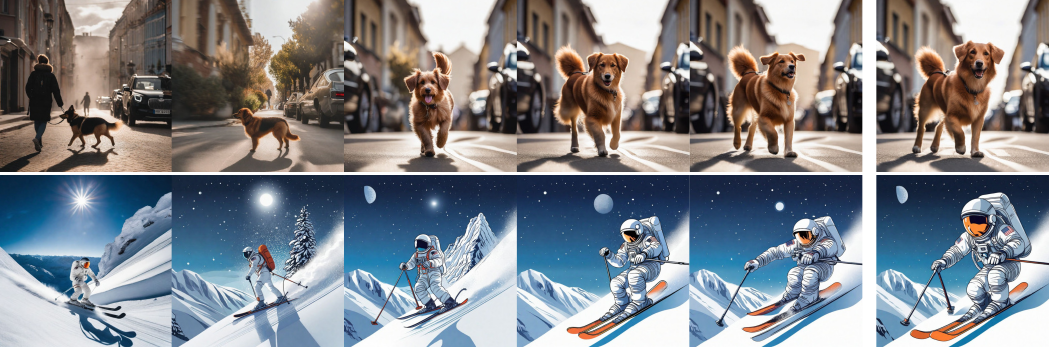}
    \vspace{-0.1cm}
    \caption{\textbf{Image Variations (Generated Images)}: The appearance similarity head can be used to create variations of a reference generated image. We show a random seed with no control (weight = 0.0) and visualize how the generation changes as we increase the strength of our Readout Guidance. Note that due to our definition of appearance similarity, identity is largely preserved while pose can change.}
    \label{fig:image_var}
    \vspace{-0.5cm}
\end{figure*}
\begin{figure*}
    \centering
    \includegraphics[width=0.98\linewidth]{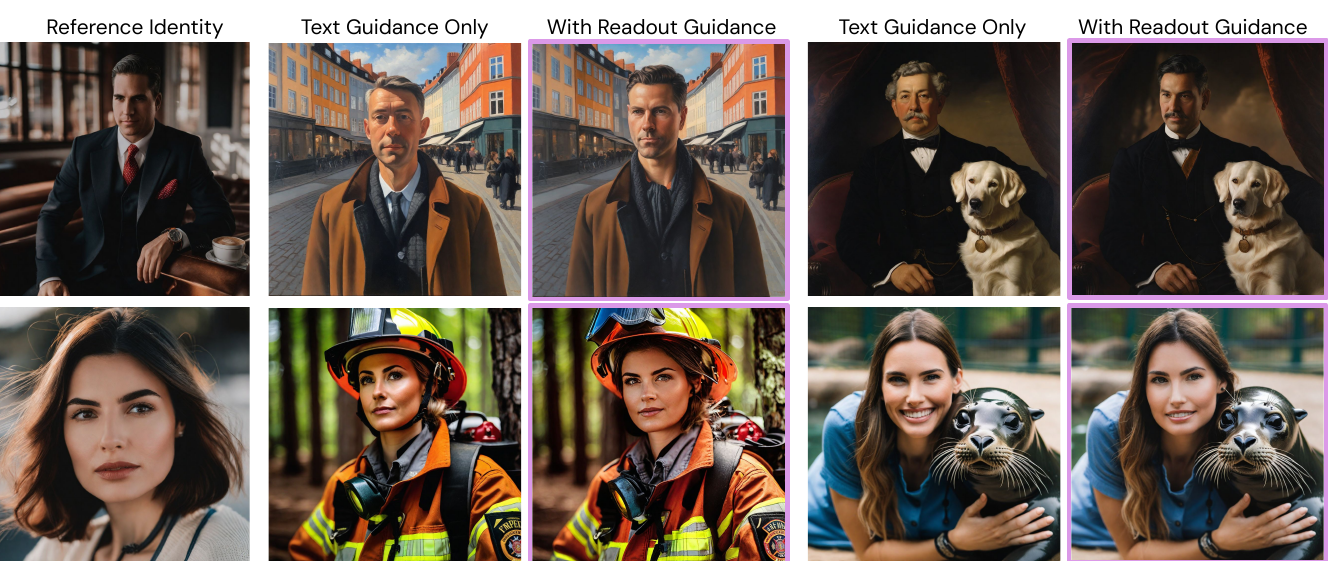}
    \caption{\textbf{Identity Guidance (Generated Images)}: Given a reference image (left), a specialized appearance similarity head can be used to guide a generated image with a different prompt to match the identity of the person in the reference image.}
    \label{fig:identity}
    \vspace{-0.3cm}
\end{figure*}
\\\textbf{Appearance Preservation.} An important property of drag-based manipulation approaches is the ability to preserve subject identity or appearance. This task is related to existing work in model personalization~\cite{ruiz2023dreambooth, ruiz2023hyperdreambooth, chen2023suti, epstein2023selfguidance, tang2023realfill}, which typically requires model fine-tuning to learn the appearance of a given subject from multiple input observations. Our appearance heads can be used for a similar application, but without subject-specific fine-tuning on a reference image. Instead, we only apply guidance against a reference readout at sampling time, allowing our method to transform an image from a random noise seed to have consistent appearance with a reference image. In~\autoref{fig:image_var} we ablate the effect of changing the guidance weight of our appearance similarity head for the same random seed. Since the notion of consistent appearance is somewhat ill-defined, by varying the weight of our guidance one can explore different definitions of this property, from only shared color and texture (columns 2, 3) to shared object identity and proportion (columns 4, 5).\\
\noindent\textbf{Identity Consistency.} The idea of appearance preservation can be further applied to specialized domains, for example people, where generative models often struggle to maintain identity across multiple generations. We showcase this potential application in~\autoref{fig:identity}, where we show that our identity head can be used to encourage generated images to contain the same person as the reference image, effectively enabling the insertion of a consistent identity into different contexts. Additional details on this guidance procedure are provided in the Supplemental. %
\\
\noindent\textbf{Spatially Aligned Control.} We also demonstrate that our spatial heads can be used to handle commonly-used control signals, including pose, depth, and edge inputs. We take images from the unseen MSCOCO~\cite{lin2014microsoft} validation set and extract their associated controls from pretrained models~\cite{Ranftl2022, 8765346, xie2015holistically}. 
\autoref{fig:controlnet} shows qualitative examples of our synthesized images and their associated readouts used during guidance. 
In~\autoref{fig:controlnet_ours}, we demonstrate that our guidance method is complementary to existing methods for conditional control. In fact, we can use the same guidance head trained on diffusion features from the vanilla base model on a model augmented with ControlNet or T2IAdapter.
When combining approaches, we note that while ControlNet~\cite{zhang2023adding} and T2I-Adapter~\cite{mou2023t2i} work well for the task of pose control, they sometimes exhibit last-mile artifacts where the synthesized pose is slightly incorrect, and our guidance is able to refine these mistakes in those cases.
\autoref{fig:train_size} also demonstrates that our method is competitive even when trained on limited data---as few as 100 training examples. We ablate training our pose readout head on 100, 1k, and all 8.5k images from PascalVOC~\cite{pascal-voc-2012} and demonstrate that all variants can handle challenging cases such as occlusions and multi-person control.
\autoref{fig:pose_plot} provides a quantitative comparison of the input poses against OpenPose's~\cite{8765346} pose prediction of our generated image by computing the percentage of correct keypoints (PCK) below a certain error threshold. We normalize the PCK threshold by the size of the predicted pose, setting it to a constant scalar $\alpha$ multiplied by the size of the predicted pose bounding box. %
We observe that training on more data generally improves quality, where our best performing method is the one trained on all 8.5k images from PascalVOC~\cite{pascal-voc-2012}. Interestingly, as seen in~\autoref{fig:pose_plot} our method trained on this relatively small dataset performs comparably to T2I-Adapter~\cite{mou2023t2i}, which was trained on 3M images from LAION-Aesthetics V2~\cite{t2iadapterpose}. Moreover, combining T2I-Adapter + Readout Guidance significantly improves the performance of the base conditional model by 30\% PCK ($\alpha=0.05$), or 2.3x in PCK.


\subsection{Limitations}
Our method has a few limitations commonly associated with sampling-time guidance: 
(1) our method generally requires more memory and runtime during sampling to compute a gradient through the intermediate features via backpropagation, even when using the same number of sampling steps as the baselines; (2) our method can sometimes produce improbable images, in the form of cartoonish or unrealistic imagery satisfying the readout constraints. We provide additional discussion in the Supplemental.

\section{Conclusion}
We have introduced \methodname{}, an approach to control the sampling of pre-trained diffusion models using readout heads trained on the diffusion features. 
Our method requires only small scale training data, making it easy to add to existing models. Because our method is a guidance technique, it can be used with both vanilla diffusion models as well as fine-tuned conditional models~\cite{zhang2023adding, mou2023t2i}. We show that the general nature of~\methodname{} allows our approach to be used for a diverse set of controls, including point correspondences and appearance similarity.

\section{Acknowledgements}
We would like to thank Stephanie Fu, Eric Wallace, Yossi Gandelsman, Dave Epstein, Ben Poole, Thomas Iljic, Brent Yi, Kevin Black, Jessica Dai, Ethan Weber, Rundi Wu, Xiaojuan Wang, Luming Tang, Ruiqi Gao, Jason Baldridge, Zhengqi Li, and Angjoo Kanazawa for helpful discussions and feedback.

\raggedbottom
\pagebreak
{
    \small
    \bibliographystyle{ieeenat_fullname}
    \bibliography{main}

\begin{thebibliography}{71}
\providecommand{\natexlab}[1]{#1}
\providecommand{\url}[1]{\texttt{#1}}
\expandafter\ifx\csname urlstyle\endcsname\relax
  \providecommand{\doi}[1]{doi: #1}\else
  \providecommand{\doi}{doi: \begingroup \urlstyle{rm}\Url}\fi

\bibitem[alp()]{alpineibex}
\url{https://www.pexels.com/video/an-alpine-ibex-with-horns-8217800/}.

\bibitem[wol()]{wolflooking}
\url{https://www.pexels.com/video/wolf-looking-around-10727436/}.

\bibitem[AI(2022)]{sddepth}
Stability AI.
\newblock Stable {D}iffusion {D}epth.
\newblock \url{https://huggingface.co/stabilityai/stable-diffusion-2-depth}, 2022.

\bibitem[Bansal et~al.(2023)Bansal, Chu, Schwarzschild, Sengupta, Goldblum, Geiping, and Goldstein]{bansal2023universal}
Arpit Bansal, Hong-Min Chu, Avi Schwarzschild, Soumyadip Sengupta, Micah Goldblum, Jonas Geiping, and Tom Goldstein.
\newblock Universal {G}uidance for {D}iffusion {M}odels.
\newblock In \emph{CVPR}, pages 843--852, 2023.

\bibitem[Baranchuk et~al.(2022)Baranchuk, Rubachev, Voynov, Khrulkov, and Babenko]{baranchuk2021label}
Dmitry Baranchuk, Ivan Rubachev, Andrey Voynov, Valentin Khrulkov, and Artem Babenko.
\newblock {Label-Efficient Semantic Segmentation with Diffusion Models}.
\newblock In \emph{ICLR}, 2022.

\bibitem[Brooks et~al.(2023)Brooks, Holynski, and Efros]{brooks2023instructpix2pix}
Tim Brooks, Aleksander Holynski, and Alexei~A Efros.
\newblock {InstructPix2Pix: Learning to Follow Image Editing Instructions}.
\newblock In \emph{CVPR}, pages 18392--18402, 2023.

\bibitem[Cao et~al.(2023)Cao, Wang, Qi, Shan, Qie, and Zheng]{cao2023masactrl}
Mingdeng Cao, Xintao Wang, Zhongang Qi, Ying Shan, Xiaohu Qie, and Yinqiang Zheng.
\newblock {MasaCtrl: Tuning-Free Mutual Self-Attention Control for Consistent Image Synthesis and Editing}.
\newblock In \emph{ICCV}, 2023.

\bibitem[{Cao} et~al.(2019){Cao}, {Hidalgo Martinez}, {Simon}, {Wei}, and {Sheikh}]{8765346}
Z. {Cao}, G. {Hidalgo Martinez}, T. {Simon}, S. {Wei}, and Y.~A. {Sheikh}.
\newblock {OpenPose: Realtime Multi-Person 2D Pose Estimation using Part Affinity Fields}.
\newblock \emph{IEEE Transactions on Pattern Analysis and Machine Intelligence}, 2019.

\bibitem[Chechik et~al.(2010)Chechik, Sharma, Shalit, , and Bengio]{gal2010largescale}
Gal Chechik, Varun Sharma, Uri Shalit, , and Samy Bengio.
\newblock {Large Scale Online Learning of Image Similarity Through Ranking}.
\newblock In \emph{JMLR}, 2010.

\bibitem[Chen et~al.(2023{\natexlab{a}})Chen, Laina, and Vedaldi]{chen2023training}
Minghao Chen, Iro Laina, and Andrea Vedaldi.
\newblock {Training-Free Layout Control with Cross-Attention Guidance}.
\newblock \emph{arXiv preprint arXiv:2304.03373}, 2023{\natexlab{a}}.

\bibitem[Chen et~al.(2023{\natexlab{b}})Chen, Hu, Li, Ruiz, Jia, Chang, and Cohen]{chen2023suti}
Wenhu Chen, Hexiang Hu, Yandong Li, Nataniel Ruiz, Xuhui Jia, Ming-Wei Chang, and William~W Cohen.
\newblock {Subject-driven Text-to-Image Generation via Apprenticeship Learning}.
\newblock \emph{arXiv preprint arXiv:2304.00186}, 2023{\natexlab{b}}.

\bibitem[Deng et~al.(2009)Deng, Dong, Socher, Li, Li, and Fei-Fei]{deng2009imagenet}
Jia Deng, Wei Dong, Richard Socher, Li-Jia Li, Kai Li, and Li Fei-Fei.
\newblock {ImageNet: A Large-Scale Hierarchical Image Database}.
\newblock In \emph{CVPR}, pages 248--255. Ieee, 2009.

\bibitem[Dhariwal and Nichol(2021)]{dhariwal2021diffusion}
Prafulla Dhariwal and Alexander Nichol.
\newblock {Diffusion Models Beat GANs on Image Synthesis}.
\newblock In \emph{NeurIPS}, pages 8780--8794, 2021.

\bibitem[Diffusers(2023)]{sdimg2img}
Diffusers.
\newblock {Image-to-Image}.
\newblock \url{https://huggingface.co/docs/diffusers/api/pipelines/stable_diffusion/img2img}, 2023.

\bibitem[Epstein et~al.(2023)Epstein, Jabri, Poole, Efros, and Holynski]{epstein2023selfguidance}
Dave Epstein, Allan Jabri, Ben Poole, Alexei~A. Efros, and Aleksander Holynski.
\newblock {Diffusion Self-Guidance for Controllable Image Generation}.
\newblock In \emph{NeurIPS}, 2023.

\bibitem[Everingham et~al.()Everingham, Van~Gool, Williams, Winn, and Zisserman]{pascal-voc-2012}
M. Everingham, L. Van~Gool, C.~K.~I. Williams, J. Winn, and A. Zisserman.
\newblock The {PASCAL} {V}isual {O}bject {C}lasses {C}hallenge 2012 {(VOC2012)} {R}esults.
\newblock http://www.pascal-network.org/challenges/VOC/voc2012/workshop/index.html.

\bibitem[Fu* et~al.(2023)Fu*, Tamir*, Sundaram*, Chai, Zhang, Dekel, and Isola]{fu2023learning}
Stephanie Fu*, Netanel Tamir*, Shobhita Sundaram*, Lucy Chai, Richard Zhang, Tali Dekel, and Phillip Isola.
\newblock {DreamSim: Learning New Dimensions of Human Visual Similarity using Synthetic Data}.
\newblock In \emph{NeurIPS}, 2023.

\bibitem[Geyer et~al.(2023)Geyer, Bar-Tal, Bagon, and Dekel]{geyer2023tokenflow}
Michal Geyer, Omer Bar-Tal, Shai Bagon, and Tali Dekel.
\newblock {TokenFlow: Consistent Diffusion Features for Consistent Video Editing}, 2023.

\bibitem[He et~al.(2016)He, Zhang, Ren, and Sun]{he2016deep}
Kaiming He, Xiangyu Zhang, Shaoqing Ren, and Jian Sun.
\newblock {Deep Residual Learning for Image Recognition}.
\newblock In \emph{CVPR}, pages 770--778, 2016.

\bibitem[Hedlin et~al.(2023)Hedlin, Sharma, Mahajan, Isack, Kar, Tagliasacchi, and Yi]{hedlin2023unsupervised}
Eric Hedlin, Gopal Sharma, Shweta Mahajan, Hossam Isack, Abhishek Kar, Andrea Tagliasacchi, and Kwang~Moo Yi.
\newblock {Unsupervised Semantic Correspondence Using Stable Diffusion}.
\newblock In \emph{NeurIPS}, 2023.

\bibitem[Ho and Salimans(2021)]{ho2021classifierfree}
Jonathan Ho and Tim Salimans.
\newblock {Classifier-Free Diffusion Guidance}.
\newblock In \emph{NeurIPS 2021 Workshop on Deep Generative Models and Downstream Applications}, 2021.

\bibitem[Hu et~al.(2022)Hu, Shen, Wallis, Allen-Zhu, Li, Wang, Wang, and Chen]{hu2021lora}
Edward~J Hu, Yelong Shen, Phillip Wallis, Zeyuan Allen-Zhu, Yuanzhi Li, Shean Wang, Lu Wang, and Weizhu Chen.
\newblock {LoRA: Low-Rank Adaptation of Large Language Models}.
\newblock 2022.

\bibitem[Huang et~al.(2023)Huang, Chen, Liu, Yujun, Zhao, and Jingren]{lhhuang2023composer}
Lianghua Huang, Di Chen, Yu Liu, Shen Yujun, Deli Zhao, and Zhou Jingren.
\newblock {Composer: Creative and Controllable Image Synthesis with Composable Conditions}.
\newblock 2023.

\bibitem[Huggingface(2023)]{sdresnet}
Huggingface.
\newblock {ResNetBlock2D}.
\newblock \url{https://github.com/huggingface/diffusers/blob/7457aa67cb5c75132c38507080697b7cc7c4d9e6/src/diffusers/models/resnet.py\#L746}, 2023.

\bibitem[Jayasumana et~al.(2023)Jayasumana, Ramalingam, Veit, Glasner, Chakrabarti, and Kumar]{jayasumana2023rethinking}
Sadeep Jayasumana, Srikumar Ramalingam, Andreas Veit, Daniel Glasner, Ayan Chakrabarti, and Sanjiv Kumar.
\newblock {Rethinking FID: Towards a Better Evaluation Metric for Image Generation}.
\newblock \emph{arXiv:401.09603}, 2023.

\bibitem[Karaev et~al.(2023)Karaev, Rocco, Graham, Neverova, Vedaldi, and Rupprecht]{karaev2023cotracker}
Nikita Karaev, Ignacio Rocco, Benjamin Graham, Natalia Neverova, Andrea Vedaldi, and Christian Rupprecht.
\newblock {CoTracker: It is Better to Track Together}.
\newblock \emph{arXiv:2307.07635}, 2023.

\bibitem[Karras et~al.(2018)Karras, Aila, Laine, and Lehtinen]{karras2017progressive}
Tero Karras, Timo Aila, Samuli Laine, and Jaakko Lehtinen.
\newblock {Progressive Growing of GANs for Improved Quality, Stability, and Variation}.
\newblock In \emph{ICLR}, 2018.

\bibitem[Khachatryan et~al.(2023)Khachatryan, Movsisyan, Tadevosyan, Henschel, Wang, Navasardyan, and Shi]{khachatryan2023text2video}
Levon Khachatryan, Andranik Movsisyan, Vahram Tadevosyan, Roberto Henschel, Zhangyang Wang, Shant Navasardyan, and Humphrey Shi.
\newblock {Text2Video-Zero: Text-to-Image Diffusion Models are Zero-Shot Video Generators}.
\newblock In \emph{ICCV}, 2023.

\bibitem[Kim et~al.(2022)Kim, Jang, Lee, Hong, Seo, and Kim]{kim2022dag}
Gyeongnyeon Kim, Wooseok Jang, Gyuseong Lee, Susung Hong, Junyoung Seo, and Seungryong Kim.
\newblock {DAG: Depth-Aware Guidance with Denoising Diffusion Probabilistic Models}.
\newblock \emph{arXiv preprint arXiv: Arxiv-2212.08861}, 2022.

\bibitem[Kingma and Ba(2015)]{kingma2014adam}
Diederik~P Kingma and Jimmy Ba.
\newblock {Adam: A Method for Stochastic Optimization}.
\newblock In \emph{ICLR}, 2015.

\bibitem[Lab(2023)]{ipadapterface}
Tencent~AI Lab.
\newblock {IP Adapter SDXL Plus Face Demo}.
\newblock \url{https://github.com/tencent-ailab/IP-Adapter/blob/main/ip_adapter_sdxl_plus-face_demo.ipynb}, 2023.

\bibitem[Labs(2022)]{sdimagevar}
Lambda Labs.
\newblock {Stable Diffusion Image Variations}.
\newblock \url{https://huggingface.co/lambdalabs/sd-image-variations-diffusers}, 2022.

\bibitem[Lian et~al.(2023)Lian, Li, Yala, and Darrell]{lian2023llmgrounded}
Long Lian, Boyi Li, Adam Yala, and Trevor Darrell.
\newblock {LLM-grounded Diffusion: Enhancing Prompt Understanding of Text-to-Image Diffusion Models with Large Language Models}.
\newblock \emph{arXiv preprint arXiv:2305.13655}, 2023.

\bibitem[Lin et~al.(2014)Lin, Maire, Belongie, Hays, Perona, Ramanan, Doll{\'a}r, and Zitnick]{lin2014microsoft}
Tsung-Yi Lin, Michael Maire, Serge Belongie, James Hays, Pietro Perona, Deva Ramanan, Piotr Doll{\'a}r, and C~Lawrence Zitnick.
\newblock {Microsoft COCO: Common Objects in Context}.
\newblock In \emph{ECCV}, pages 740--755. Springer, 2014.

\bibitem[Liu et~al.(2023)Liu, Wu, Van~Hoorick, Tokmakov, Zakharov, and Vondrick]{liu2023zero}
Ruoshi Liu, Rundi Wu, Basile Van~Hoorick, Pavel Tokmakov, Sergey Zakharov, and Carl Vondrick.
\newblock {Zero-1-to-3: Zero-shot One Image to 3D Object}.
\newblock In \emph{ICCV}, pages 9298--9309, 2023.

\bibitem[Luo et~al.(2023)Luo, Dunlap, Park, Holynski, and Darrell]{luo2023dhf}
Grace Luo, Lisa Dunlap, Dong~Huk Park, Aleksander Holynski, and Trevor Darrell.
\newblock {Diffusion Hyperfeatures: Searching Through Time and Space for Semantic Correspondence}.
\newblock In \emph{NeurIPS}, 2023.

\bibitem[Meng et~al.(2022)Meng, He, Song, Song, Wu, Zhu, and Ermon]{meng2022sdedit}
Chenlin Meng, Yutong He, Yang Song, Jiaming Song, Jiajun Wu, Jun-Yan Zhu, and Stefano Ermon.
\newblock {SDEdit: Guided Image Synthesis and Editing with Stochastic Differential Equations}.
\newblock In \emph{ICLR}, 2022.

\bibitem[ML(2022)]{sdinpaint}
Runway ML.
\newblock {Stable Diffusion Inpainting}.
\newblock \url{https://huggingface.co/runwayml/stable-diffusion-inpainting}, 2022.

\bibitem[Mou et~al.(2023{\natexlab{a}})Mou, Wang, Song, Shan, and Zhang]{mou2023dragondiffusion}
Chong Mou, Xintao Wang, Jiechong Song, Ying Shan, and Jian Zhang.
\newblock {DragonDiffusion: Enabling Drag-style Manipulation on Diffusion Models}.
\newblock \emph{arXiv preprint arXiv:2307.02421}, 2023{\natexlab{a}}.

\bibitem[Mou et~al.(2023{\natexlab{b}})Mou, Wang, Xie, Zhang, Qi, Shan, and Qie]{mou2023t2i}
Chong Mou, Xintao Wang, Liangbin Xie, Jian Zhang, Zhongang Qi, Ying Shan, and Xiaohu Qie.
\newblock {T2I-Adapter: Learning Adapters to Dig out More Controllable Ability for Text-to-Image Diffusion Models}.
\newblock \emph{arXiv preprint arXiv:2302.08453}, 2023{\natexlab{b}}.

\bibitem[Org(2020)]{midasnet}
Intelligent Systems~Lab Org.
\newblock {MidasNet}.
\newblock \url{https://github.com/isl-org/MiDaS/blob/bdc4ed64c095e026dc0a2f17cabb14d58263decb/midas/midas_net.py\#L37C11-L44C10}, 2020.

\bibitem[Pan et~al.(2023)Pan, Tewari, Leimk{\"u}hler, Liu, Meka, and Theobalt]{pan2023drag}
Xingang Pan, Ayush Tewari, Thomas Leimk{\"u}hler, Lingjie Liu, Abhimitra Meka, and Christian Theobalt.
\newblock {Drag Your {GAN}: Interactive Point-based Manipulation on the Generative Image Manifold}.
\newblock In \emph{ACM SIGGRAPH 2023 Conference Proceedings}, pages 1--11, 2023.

\bibitem[Park* et~al.(2024)Park*, Luo*, Toste, Azadi, Liu, Karalashvili, Rohrbach, and Darrell]{park2024shape}
Dong~Huk Park*, Grace Luo*, Clayton Toste, Samaneh Azadi, Xihui Liu, Maka Karalashvili, Anna Rohrbach, and Trevor Darrell.
\newblock {Shape-Guided Diffusion with Inside-Outside Attention}.
\newblock In \emph{WACV}, 2024.

\bibitem[Podell et~al.(2023)Podell, English, Lacey, Blattmann, Dockhorn, Müller, Penna, and Rombach]{podell2023sdxl}
Dustin Podell, Zion English, Kyle Lacey, Andreas Blattmann, Tim Dockhorn, Jonas Müller, Joe Penna, and Robin Rombach.
\newblock {SDXL: Improving Latent Diffusion Models for High-Resolution Image Synthesis}, 2023.

\bibitem[Pont{-}Tuset et~al.(2017)Pont{-}Tuset, Perazzi, Caelles, Arbelaez, Sorkine{-}Hornung, and Gool]{DBLP:journals/corr/Pont-TusetPCASG17}
Jordi Pont{-}Tuset, Federico Perazzi, Sergi Caelles, Pablo Arbelaez, Alexander Sorkine{-}Hornung, and Luc~Van Gool.
\newblock {The 2017 {DAVIS} Challenge on Video Object Segmentation}.
\newblock \emph{CoRR}, abs/1704.00675, 2017.

\bibitem[Radford et~al.(2021{\natexlab{a}})Radford, Kim, Hallacy, Ramesh, Goh, Agarwal, Sastry, Askell, Mishkin, Clark, Krueger, and Sutskever]{clip}
Alec Radford, Jong~Wook Kim, Chris Hallacy, A. Ramesh, Gabriel Goh, Sandhini Agarwal, Girish Sastry, Amanda Askell, Pamela Mishkin, Jack Clark, Gretchen Krueger, and Ilya Sutskever.
\newblock {Learning Transferable Visual Models From Natural Language Supervision}.
\newblock In \emph{ICML}, 2021{\natexlab{a}}.

\bibitem[Radford et~al.(2021{\natexlab{b}})Radford, Kim, Hallacy, Ramesh, Goh, Agarwal, Sastry, Askell, Mishkin, Clark, et~al.]{radford2021learning}
Alec Radford, Jong~Wook Kim, Chris Hallacy, Aditya Ramesh, Gabriel Goh, Sandhini Agarwal, Girish Sastry, Amanda Askell, Pamela Mishkin, Jack Clark, et~al.
\newblock {Learning Transferable Visual Models From Natural Language Supervision}.
\newblock In \emph{ICML}, pages 8748--8763. PMLR, 2021{\natexlab{b}}.

\bibitem[Ranftl et~al.(2021)Ranftl, Bochkovskiy, and Koltun]{Ranftl2021}
Ren\'{e} Ranftl, Alexey Bochkovskiy, and Vladlen Koltun.
\newblock {Vision Transformers for Dense Prediction}.
\newblock \emph{ICCV}, 2021.

\bibitem[Ranftl et~al.(2022)Ranftl, Lasinger, Hafner, Schindler, and Koltun]{Ranftl2022}
Ren\'{e} Ranftl, Katrin Lasinger, David Hafner, Konrad Schindler, and Vladlen Koltun.
\newblock {Towards Robust Monocular Depth Estimation: Mixing Datasets for Zero-Shot Cross-Dataset Transfer}.
\newblock \emph{IEEE Transactions on Pattern Analysis and Machine Intelligence}, 44\penalty0 (3), 2022.

\bibitem[Rombach et~al.(2022)Rombach, Blattmann, Lorenz, Esser, and Ommer]{Rombach_2022_CVPR}
Robin Rombach, Andreas Blattmann, Dominik Lorenz, Patrick Esser, and Bj\"orn Ommer.
\newblock {High-Resolution Image Synthesis With Latent Diffusion Models}.
\newblock In \emph{CVPR}, pages 10684--10695, 2022.

\bibitem[Ruiz et~al.(2023{\natexlab{a}})Ruiz, Li, Jampani, Pritch, Rubinstein, and Aberman]{ruiz2023dreambooth}
Nataniel Ruiz, Yuanzhen Li, Varun Jampani, Yael Pritch, Michael Rubinstein, and Kfir Aberman.
\newblock {DreamBooth: Fine Tuning Text-to-image Diffusion Models for Subject-Driven Generation}.
\newblock In \emph{CVPR}, pages 22500--22510, 2023{\natexlab{a}}.

\bibitem[Ruiz et~al.(2023{\natexlab{b}})Ruiz, Li, Jampani, Wei, Hou, Pritch, Wadhwa, Rubinstein, and Aberman]{ruiz2023hyperdreambooth}
Nataniel Ruiz, Yuanzhen Li, Varun Jampani, Wei Wei, Tingbo Hou, Yael Pritch, Neal Wadhwa, Michael Rubinstein, and Kfir Aberman.
\newblock {HyperDreamBooth: HyperNetworks for Fast Personalization of Text-to-Image Models}, 2023{\natexlab{b}}.

\bibitem[Shi et~al.(2023)Shi, Xue, Pan, Zhang, Tan, and Bai]{shi2023dragdiffusion}
Yujun Shi, Chuhui Xue, Jiachun Pan, Wenqing Zhang, Vincent~YF Tan, and Song Bai.
\newblock {DragDiffusion: Harnessing Diffusion Models for Interactive Point-based Image Editing}.
\newblock \emph{arXiv preprint arXiv:2306.14435}, 2023.

\bibitem[Song et~al.(2021)Song, Meng, and Ermon]{song2020denoising}
Jiaming Song, Chenlin Meng, and Stefano Ermon.
\newblock {Denoising Diffusion Implicit Models}.
\newblock In \emph{ICLR}, 2021.

\bibitem[Tang et~al.(2023{\natexlab{a}})Tang, Jia, Wang, Phoo, and Hariharan]{tang2023dift}
Luming Tang, Menglin Jia, Qianqian Wang, Cheng~Perng Phoo, and Bharath Hariharan.
\newblock {Emergent Correspondence from Image Diffusion}.
\newblock In \emph{NeurIPS}, 2023{\natexlab{a}}.

\bibitem[Tang et~al.(2023{\natexlab{b}})Tang, Ruiz, Chu, Li, Holynski, Jacobs, Hariharan, Pritch, Wadhwa, Aberman, et~al.]{tang2023realfill}
Luming Tang, Nataniel Ruiz, Qinghao Chu, Yuanzhen Li, Aleksander Holynski, David~E Jacobs, Bharath Hariharan, Yael Pritch, Neal Wadhwa, Kfir Aberman, et~al.
\newblock {RealFill: Reference-Driven Generation for Authentic Image Completion}.
\newblock \emph{arXiv preprint arXiv:2309.16668}, 2023{\natexlab{b}}.

\bibitem[TencentARC(2023)]{t2iadapterpose}
TencentARC.
\newblock {T2I-Adapter-SDXL - Openpose}.
\newblock \url{https://huggingface.co/TencentARC/t2i-adapter-openpose-sdxl-1.0}, 2023.

\bibitem[timesler(2019)]{facenetpytorch}
timesler.
\newblock facenet-pytorch.
\newblock \url{https://github.com/timesler/facenet-pytorch}, 2019.

\bibitem[Voynov et~al.(2023)Voynov, Abernan, and Cohen-Or]{voynov2022sketch}
Andrey Voynov, Kfir Abernan, and Daniel Cohen-Or.
\newblock {Sketch-Guided Text-to-Image Diffusion Models}.
\newblock In \emph{SIGGRAPH}, 2023.

\bibitem[Wallace et~al.(2023)Wallace, Gokul, Ermon, and Naik]{wallace2023endtoend}
Bram Wallace, Akash Gokul, Stefano Ermon, and Nikhil Naik.
\newblock {End-to-End Diffusion Latent Optimization Improves Classifier Guidance}.
\newblock In \emph{ICCV}, 2023.

\bibitem[Whitaker(2023)]{whitaker2023midu}
Jonathan Whitaker.
\newblock {Mid-U Guidance: Fast Classifier Guidance for Latent Diffusion Models}.
\newblock \url{https://wandb.ai/johnowhitaker/midu-guidance/reports/Mid-U-Guidance-Fast-Classifier-Guidance-for-Latent-Diffusion-Models--VmlldzozMjg0NzA1}, 2023.

\bibitem[Wu et~al.(2023{\natexlab{a}})Wu, Ge, Wang, Lei, Gu, Shi, Hsu, Shan, Qie, and Shou]{wu2023tune}
Jay~Zhangjie Wu, Yixiao Ge, Xintao Wang, Stan~Weixian Lei, Yuchao Gu, Yufei Shi, Wynne Hsu, Ying Shan, Xiaohu Qie, and Mike~Zheng Shou.
\newblock {Tune-A-Video: One-Shot Tuning of Image Diffusion Models for Text-to-Video Generation}.
\newblock In \emph{ICCV}, pages 7623--7633, 2023{\natexlab{a}}.

\bibitem[Wu et~al.(2023{\natexlab{b}})Wu, Zhao, Chen, Gu, Zhao, He, Zhou, Shou, and Shen]{wu2023datasetdm}
Weijia Wu, Yuzhong Zhao, Hao Chen, Yuchao Gu, Rui Zhao, Yefei He, Hong Zhou, Mike~Zheng Shou, and Chunhua Shen.
\newblock {DatasetDM: Synthesizing Data with Perception Annotations Using Diffusion Models}.
\newblock In \emph{NeurIPS}, 2023{\natexlab{b}}.

\bibitem[Xiang and Zhu(2017)]{xiang2017joint}
Jia Xiang and Gengming Zhu.
\newblock {Joint Face Detection and Facial Expression Recognition with MTCNN}.
\newblock In \emph{ICISCE}, pages 424--427. IEEE, 2017.

\bibitem[Xie and Tu(2015)]{xie2015holistically}
Saining Xie and Zhuowen Tu.
\newblock {Holistically-Nested Edge Detection}.
\newblock In \emph{ICCV}, pages 1395--1403, 2015.

\bibitem[Xu et~al.(2023)Xu, Liu, Vahdat, Byeon, Wang, and De~Mello]{xu2022odise}
Jiarui Xu, Sifei Liu, Arash Vahdat, Wonmin Byeon, Xiaolong Wang, and Shalini De~Mello.
\newblock {ODISE: Open-Vocabulary Panoptic Segmentation with Text-to-Image Diffusion Models}.
\newblock In \emph{CVPR}, 2023.

\bibitem[Ye et~al.(2023)Ye, Zhang, Liu, Han, and Yang]{ye2023ip-adapter}
Hu Ye, Jun Zhang, Sibo Liu, Xiao Han, and Wei Yang.
\newblock {IP-Adapter: Text Compatible Image Prompt Adapter for Text-to-Image Diffusion Models}.
\newblock 2023.

\bibitem[Zhang et~al.(2023{\natexlab{a}})Zhang, Herrmann, Hur, Cabrera, Jampani, Sun, and Yang]{zhang2023tale}
Junyi Zhang, Charles Herrmann, Junhwa Hur, Luisa~Polania Cabrera, Varun Jampani, Deqing Sun, and Ming-Hsuan Yang.
\newblock {A Tale of Two Features: Stable Diffusion Complements DINO for Zero-Shot Semantic Correspondence}.
\newblock In \emph{NeurIPS}, 2023{\natexlab{a}}.

\bibitem[Zhang(2023)]{controlpose}
Lvmin Zhang.
\newblock {Controlnet - Human Pose Version}.
\newblock \url{https://huggingface.co/lllyasviel/sd-controlnet-openpose}, 2023.

\bibitem[Zhang et~al.(2023{\natexlab{b}})Zhang, Rao, and Agrawala]{zhang2023adding}
Lvmin Zhang, Anyi Rao, and Maneesh Agrawala.
\newblock {Adding Conditional Control to Text-to-Image Diffusion Models}.
\newblock In \emph{ICCV}, 2023{\natexlab{b}}.

\bibitem[Zhao et~al.(2023)Zhao, Rao, Liu, Liu, Zhou, and Lu]{zhao2023unleashing}
Wenliang Zhao, Yongming Rao, Zuyan Liu, Benlin Liu, Jie Zhou, and Jiwen Lu.
\newblock {Unleashing Text-to-Image Diffusion Models for Visual Perception}.
\newblock In \emph{ICCV}, 2023.

\end{thebibliography}
}

\clearpage
\setcounter{page}{1}
\maketitlesupplementary
\raggedbottom
\section{Readout Guidance}
\begin{table*}
\centering
\begin{small}
\begin{tabular}{l|cccccc}
\toprule
Task & text\_weight  & rg\_weight & rg\_ratio & $\eta$\\
\midrule
Drag-Based Manipulation (Real Images) & 3.5  & 1.0 & [0.0, 0.5] & 0.0 \\
Drag-Based Manipulation (Generated Images) & 7.5  & 1.0 & [0.0, 0.5] & 0.0 \\
Appearance Preservation (Generated Images) & 7.5  & 1.0 & [0.05, 0.5] & 0.0\\
Identity Consistency (Generated Images) & 7.5 & 1.0 & [0.05, 0.8] & 1.0\\
Spatially Aligned Control (Generated Images) & 7.5 & 0.5 & [0.0, 0.5] & 1.0\\
\bottomrule
\end{tabular}
\end{small}
\caption{\textbf{Guidance Hyperparameters:} We recommend the following hyperparameter settings for each task when using~\guidancename{}.}
\label{tab:hparams}
\end{table*}
\subsection{Guidance Hyperparameters}
For all tasks, we use $100$ sampling timesteps, which we find produces results that are faithful to the input control while still maintaining a reasonable runtime. Our guidance update is rescaled by $2e^{-2}$ to allow the user-defined guidance weight to occupy a more intuitive range of $[0.0, 1.0]$, where $0$ is no additional guidance, and $1$ is very strong guidance (values $>1$ are also valid, but typically not desirable). Below, we describe the set of hyperparameters that vary between applications shown in the paper. For each of these, the values used for generating our results are shown in~\autoref{tab:hparams}.\\
\textbf{text\_weight:} the weight of text classifier-free guidance~\cite{ho2021classifierfree}, which ranges between $[0, \infty]$.\\
\textbf{rg\_weight:} the weight of our guidance signal, which ranges between $[0, \infty]$. As mentioned above and demonstrated in~\autoref{fig:failure_cases_suppl}, values much larger than $1.0$ can result in visual artifacts. The effects of varying this weight in can be seen in~\autoref{fig:image_var} of the main text.\\
\textbf{rg\_ratio:} the span of timesteps [start, end] to apply our guidance to, expressed as a fraction of the total number of timesteps. These values range between $[0, 1.0]$. We find that early diffusion steps generally determine structure and composition, while late diffusion steps determine high frequency detail and photorealism. For the appearance preservation and identity consistency tasks, we find it helpful to set the start value $>0.0$ to retain structural diversity from a random seed (setting start $=0.0$ can cause the reference image's structure to be copied in addition to the subject's identity). For all tasks, we set end $<1.0$, to avoid the guidance signal competing with photorealism.\\
\textbf{eta ($\eta$):} the stochasticity of the sampling process, which ranges between [0, 1.0]. Setting $\eta=0.0$ is traditional deterministic DDIM sampling~\cite{song2020denoising} and $\eta=1.0$ is highly stochastic. Empirically we also find that using $\eta=1.0$ is especially critical for the spatially aligned control task, where using $\eta=0.0$ can result in visual artifacts, which we show in~\autoref{fig:failure_cases_suppl}. We hypothesize that stochasticity is required when regressing against a user input control, which is more out of domain compared to a reference set of diffusion features.\\
\begin{figure}
    \includegraphics[width=1.0\linewidth]{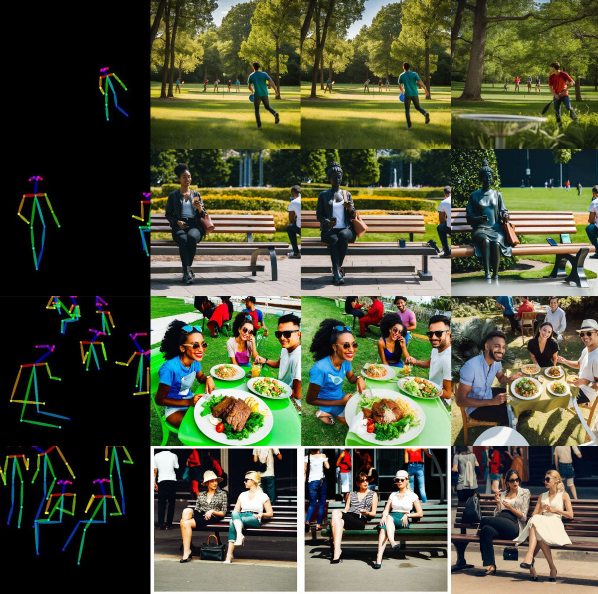}
    \hfill
    \begin{subfigure}{0.24\linewidth}
        \centering
        \caption{Pose Control}
    \end{subfigure} \hfill
    \begin{subfigure}{0.24\linewidth}
        \centering
        \caption{w/ $\tau_t$}
        \label{subfig:cg}
    \end{subfigure} \hfill
    \begin{subfigure}{0.24\linewidth}
        \centering
         \caption{w/o $\tau_t$}
         \label{subfig:no_timescale}
    \end{subfigure} \hfill
    \begin{subfigure}{0.24\linewidth}
        \centering
        \caption{Normalized}
        \label{subfig:l2norm}
    \end{subfigure}
    \hfill
    \caption{\textbf{Update Rule Comparison}: For the same pose control, we compare outputs when using timestep rescaling as was done in original classifier guidance~\cite{dhariwal2021diffusion}, removing the timestep rescaling factor, and normalizing the gradient with the L2-norm.}
    \label{fig:sgd_vs_adam}
    \vspace{0cm}
\end{figure}
\subsection{Update Rule}\label{sec:update_rule}
We observe that our guidance process is sensitive to the magnitude of the gradient signal at each sampling step. In this section, we explore different mechanisms for scaling the gradient signal, inspired by common practices in gradient descent optimization.
First, recall that our method derives from the original classifier guidance (CG) update rule~\cite{dhariwal2021diffusion}, which is written as follows:
$$\hat{\epsilon}_t \gets \epsilon_{\theta}(x_t) - \sqrt{1 - \bar{\alpha}_t} \nabla_{x_t} \mathrm{log}\, p_{\phi}(y|x_t)$$
where, in our case, we replace the gradient term $\mathrm{log}\, p_{\phi}(y|x_t)$ with a distance function over the readout $d(r, f_{\psi}(x_t))$. For simplicity, we can rewrite this update rule as a standard optimizer update:
$$w_t \gets w_{t-1} - \alpha \cdot \tau_{t} \cdot \nabla \mathcal{L},$$ where $w$ represents the input/output noise, $\tau_{t}$ is a timestep-dependent rescaling, $\alpha$ is a configurable learning rate, and $\mathcal{L}$ is the loss or distance function. Following this formulation, we explore a couple of alternative update rules for applying guidance, with qualitative comparisons shown in~\autoref{fig:sgd_vs_adam}. First, since our readout heads are already timestep-conditional, we ablate the necessity of timestep rescaling: 
$$w_t \gets w_{t-1} - \alpha \cdot \nabla \mathcal{L},$$
We use the learning rate $\alpha = 200$ for both this variant and the one with $\tau_{t}$.
We also experiment with normalizing the gradient by the L2-norm, as might be done in an Adam~\cite{kingma2014adam} update (without the historical moving average):
$$w_t \gets w_{t-1} - \alpha \cdot \frac{\nabla\mathcal{L}}{\sqrt{\nabla\mathcal{L}^2}+\epsilon},$$
where $\epsilon$ is a small positive constant to prevent division by zero. For this variant, we use $\alpha=2e^{-2}$.
Unlike the other two update rules, the normalized gradient variant has an effective step size bounded only by the learning rate $\alpha$, and is therefore invariant to the magnitude of the gradient $\nabla\mathcal{L}$~\cite{kingma2014adam}.

Empirically, we find that when using our readout heads, timestep rescaling in the update rule has a negligible effect, often producing outputs that are extremely similar to those without rescaling (\autoref{subfig:cg}~and~\autoref{subfig:no_timescale}). We further notice that making the readout heads timestep-conditional is very important---removing this conditioning results in improbable outputs that respect the input control but quickly fall off the manifold of natural imagery. A similar failure mode occurs if the magnitude of the loss $\nabla\mathcal{L}$ is too large. For this reason, we find that the normalized gradient update rule (\autoref{subfig:l2norm}) produces the highest fidelity results, as it causes the update step size to be agnostic to gradient scale. Note the difference between Figure~\ref{subfig:no_timescale}~and~\ref{subfig:l2norm}, where the un-normalized update rules sometimes exhibit contrast and saturation artifacts. All the results in our paper use the normalized variant, although our method is also effective with the other two update rules. %
\begin{figure*}
    \includegraphics[trim={0cm 4.5cm 9.5cm 0cm},clip,width=1.0\linewidth]{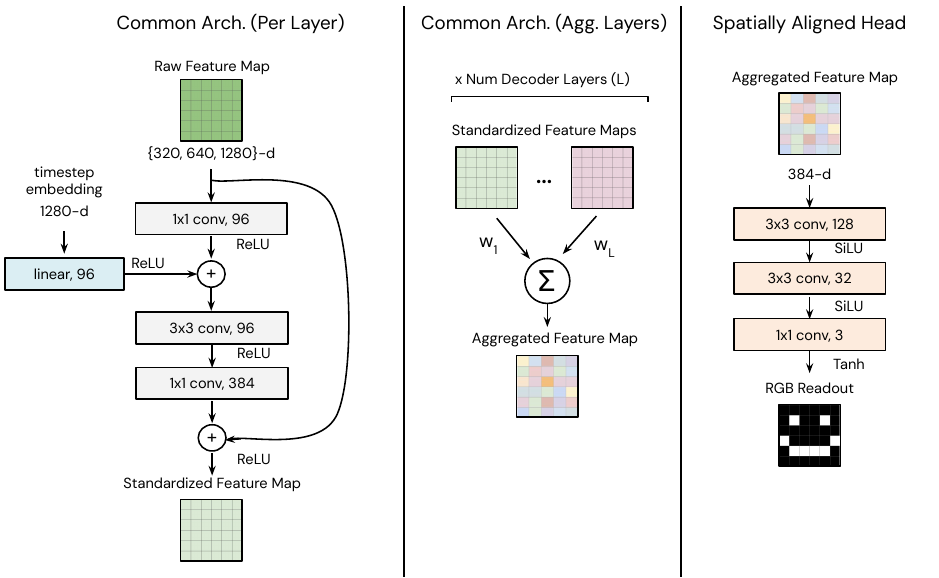} 
    \vspace{-0.3cm}
    \caption{\textbf{Architecture Diagram}: We share a common architecture 
 across all readout heads, where we first convert the raw feature map to a feature map of a standardized channel count and spatial resolution (left), aggregate feature maps from all decoder layers (middle), and optionally convert the aggregated feature map into an RGB readout (right).}
    \label{fig:architecture_diagram}
    \vspace{-0.3cm}
\end{figure*}
\subsection{Sampling Compute}
Here we report the runtime and memory consumption on an Nvidia A100 GPU.
For the pose control task, sampling with Readout Guidance takes 32.7s on SDv1-5 (vs. 10.7s with only text guidance) and consumes 11.9GB of VRAM (vs. 8.5GB). On SDXL, sampling with Readout Guidance takes 44.0s (vs. 14.2s) and consumes 27.7GB of VRAM (vs. 15.6GB). For the drag-based manipulation task, sampling with Readout Guidance takes 55.4s (vs. 16.6s) on SDv1-5 and consumes 19.2GB of VRAM (vs. 8.3GB). On SDXL, Readout Guidance takes 75.0s (vs. 23.4s) and consumes 37.6GB of VRAM (vs. 15.8GB). 

It is important to note that we use a naive initial implementation that has not been optimized for memory usage---\ie, we retain all relevant buffers in memory: all features, gradients, target control signals. A number of optimizations would likely reduce memory consumption to less than half the reported numbers, \eg, not batching the conditional and unconditional inputs together, pre-computing and caching reference features for pairwise tasks, among others. Using the current implementation, the most memory-efficient setting is able to run on readily accessible GPUs such as an Nvidia T4 (16GB VRAM) available via Google Colab, and we anticipate further memory optimization should enable our method's use on even lower-end GPUs.

\subsection{Training Compute} Here we share the training time for each readout head on an Nvidia A100 GPU. We train the pose head on PascalVOC with a batch size of 8 image / control pairs for at most 5k steps, which takes 37 min for SDv1-5 and 2 hr 25 min for SDXL. The depth head takes 41 min for SDv1-5 and 2 hr 40 min for SDXL. The edge head takes 41 min for SDv1-5 and 2 hr 38 min for SDXL. We train the appearance similarity head on DAVIS with a batch size of 1 anchor / positive / negative triplet for at most 1k steps, which takes 08 min for SDv1-5 and 24 min for SDXL. We train the correspondence feature head on DAVIS with a batch size of 1 frame / frame pair for at most 10k steps, which takes 48 min for SDv1-5 and 2 hr 15 min for SDXL.

\subsection{Readout Head Architecture} In~\autoref{fig:architecture_diagram} we depict a detailed architecture diagram of our readout heads. We share a common architecture for all readout heads, building off of the aggregation network proposed in Diffusion Hyperfeatures~\cite{luo2023dhf}. For each raw decoder feature map, Diffusion Hyperfeatures uses bottleneck layers~\cite{he2016deep} to standardize the channel count (left column), then aggregates these standardized feature maps with a learned weighted sum (middle column). Our main modification to adapt this architecture for guidance is to make the bottleneck layers timestep-conditional, implemented as the blue linear layer shown in the left column of~\autoref{fig:architecture_diagram}. More specifically, we use the U-Net's pre-trained timestep embedding layers to get a timestep embedding, learn a projection layer, and add this projected embedding to the feature map during the standardization process. 
Our design was inspired by the Stable Diffusion implementation of timestep conditioning in their ResNet blocks~\cite{sdresnet}. 
For the case of the spatially aligned heads (right column), we make another modification where we add convolutional layers (yellow) that convert the multi-channel aggregated feature map into a three-channel RGB image. Our design was inspired by the output convolutions used in MiDaS~\cite{midasnet, Ranftl2022} and the SiLU (a soft version of ReLU) activations used in ControlNet~\cite{zhang2023adding}. We use Tanh for our final activation function to ensure that the output is bounded between $[-1, 1]$, rather than unbounded.

\subsection{Improbable Images} In~\autoref{fig:failure_cases_suppl} we show a few failure cases. Although these images fully respect the input pose control, they start to diverge from the natural image manifold, tending towards cartoonish colors or surreal imagery with relatively texture-less scenes. 
These types of artifacts are likely the result of a trade-off between the guidance target and photorealism (as defined by the base model's log-likelihood)---and can be resolved by lowering the readout guidance weight, stopping our guidance earlier in the diffusion process, or starting from a different initial seed.

\anycolfigpdf{3}{figures}{
  improbable_images.pdf/1/Pose Control,
  improbable_images.pdf/2/Pose Control,
  improbable_images.pdf/3/Pose Control
}{Improbable Image}{fig:failure_cases_suppl}{\textbf{Improbable Images}: We show images generated with our pose head using a deterministic sampler ($\eta=0.0$), rather than the recommended stochastic sampler ($\eta=1.0$). These types of generations also arise when the guidance weight is set too high.}{0.8cm}{-0.5cm}

\begin{table*}[h]
\centering
\begin{small}
\setlength{\tabcolsep}{3pt} %
\begin{tabular}{l@{\quad}|c|c|c@{\quad}c@{\quad}c}
\toprule
Method & Added Training Data & Added Params / Size & PCK$_{\alpha=0.05}$ ($\uparrow$) & PCK$_{\alpha=0.1}$ ($\uparrow$) & PCK$_{\alpha=0.2}$ ($\uparrow$)\\
\midrule
\textbf{SDv1-5} & & & & \\
No Control & - & - & 0.49 & 1.55 & 5.42  \\
RG$_{\text{8.5k}}$ & 8.5k Images & 8.5M / 49MB & 19.91 & 36.10 & 50.71\\
ControlNet~\cite{zhang2023adding} & 200k Images & 361M / 1.4GB & 42.34 & 60.22 & 73.38\\
ControlNet + RG$_{\text{8.5k}}$  & 200k + 8.5k Images & 370M / 1.4GB & \textbf{55.82} & \textbf{77.47} & \textbf{86.93}\\
\midrule
\textbf{SDXL} & & & & \\
No Control & - & - & 1.19 & 2.94 & 10.40\\
RG$_{\text{100}}$ & 100 Images & 5.9M / 35MB & 15.24 & 28.19 & 41.62\\
RG$_{\text{1k}}$ & 1k Images & 5.9M / 35MB & 19.65 & 29.41 & 43.17\\
RG$_{\text{8.5k}}$ & 8.5k Images & 5.9M / 35MB & 24.29 & 37.07 & 46.21\\
T2IAdapter~\cite{mou2023t2i} & 3M Images & 79M / 302MB & 24.06 & 35.92 & 49.49\\
\text{T2IAdapter}+RG$_{\text{8.5k}}$ & 3M + 8.5k Images & 85M / 337MB & \textbf{54.54} & \textbf{72.04} & \textbf{82.18}\\
\bottomrule
\end{tabular}
\end{small}
\caption{\textbf{Pose Control Comparison:} We compute the percentage of correct keypoints (PCK) at varying thresholds $\alpha$ between the input pose and the pose of the generated image on 100 random images of humans from the MSCOCO~\cite{lin2014microsoft} validation set.}
\label{tab:pose_pck}
\end{table*}

\begin{table}[h]
\centering
\begin{small}
\setlength{\tabcolsep}{3pt} %
\begin{tabular}{l@{\quad}|c}
\toprule
Method & CMMD ($\downarrow$)\\
\midrule
\textbf{SDXL} & \\
No Control & 1.374\\
RG$_{100}$ & 1.305\\
RG$_{1k}$ & 1.243\\
RG$_{8.5k}$ & 1.134\\
T2IAdapter & 0.817\\
T2IAdapter + RG$_{8.5k}$ & 0.766\\
\bottomrule
\end{tabular}
\end{small}
\caption{\textbf{Pose Control Image Quality:} We also compute the CLIP Maximum Mean Discrepancy (CMMD)~\cite{jayasumana2023rethinking} between the distribution of real images and generated images on the same set of 100 MSCOCO images described in~\autoref{tab:pose_pck}.}
\label{tab:pose_cmmd}
\end{table}

\begin{figure*}
    \includegraphics[width=1.0\linewidth]{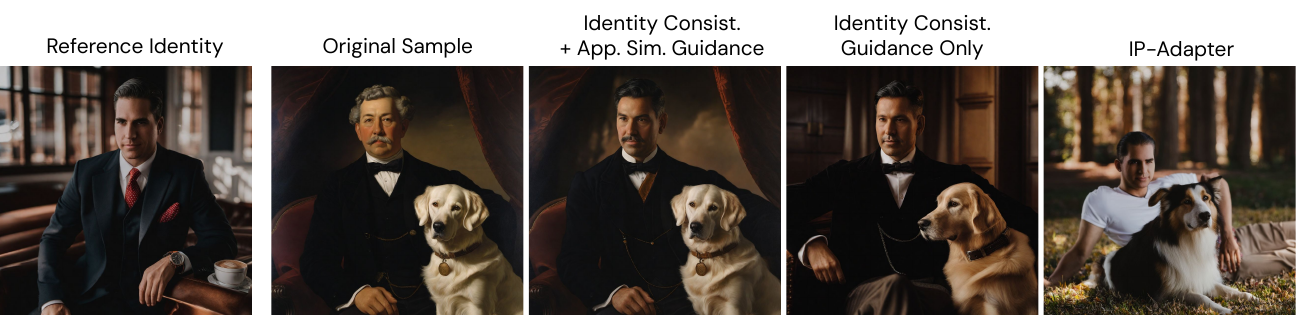} 
    \caption{\textbf{Identity Consistency Ablation}: Our full method uses the identity consistency head on the face and appearance similarity head to preserve the background. We also ablate using only the identity consistency head. Concurrent work IP-Adapter~\cite{ye2023ip-adapter} does not support injecting identity into a specific sampled image.}
    \label{fig:identity_suppl}
    \vspace{-0.3cm}
\end{figure*}

\section{Relative Control}

\subsection{Identity Consistency}
In this task, we seek to generate an image (or set of images) of the same person or character. Given an input image, a sampling process (otherwise guided by a text prompt describing a different scene) is encouraged to produce an image that contains a person with the same identity as the input image. To accomplish this, we use a specialized version of the appearance head trained on a dataset of face photographs~\cite{karras2017progressive}. Since this dataset consists of mostly tight crops around the face, we apply our guidance constraint only at pixels which we determine to belong to a face (\ie, using an off-the-shelf face detector~\cite{xiang2017joint, facenetpytorch}). While we only apply guidance within the face region, this can cause the overall sampling trajectory to change, which can in turn change the appearance of the background. To counteract this, we also apply our (standard, \ie, not identity-specialized) appearance guidance on the rest of the image, but towards the original sampling trajectory, allowing the background to remain unchanged. In~\autoref{fig:identity_suppl} we ablate the effect of removing this appearance similarity guidance (col.~4), showing that the background drifts more in appearance but often the identity is more closely matched. We also compare to the result of the concurrent work IP-Adapter~\cite{ye2023ip-adapter, ipadapterface} (col.~5), which trains an adapter to condition on the CLIP image embedding~\cite{radford2021learning} of a reference image. Although IP-Adapter can also be used to borrow a reference identity (face), unlike our method, it cannot inject this identity into a particular sampled image. 

\section{Spatially Aligned Control}
\subsection{Readout Evolution}
\begin{figure*}
    \caption*{\equallyspacedcaption{t=0}{t=10, t=20, t=30, t=40, t=100}{1cm}{-1cm}}
    \vspace{-0.3cm}
    \includegraphics[width=1.0\linewidth]{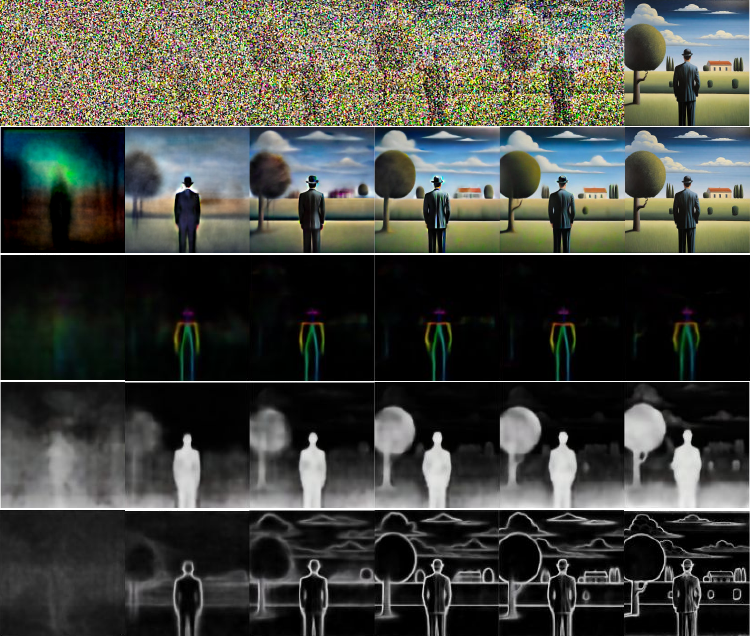} 
    \vspace{-0.3cm}
    \caption{\textbf{Readout Evolution}: We synthesize an image with the prompt \promptcaption{painting of man standing outdoors in the style of magritte}, and show (from top to bottom) the $x_t$, predicted clean image~\cite{song2020denoising}, pose readout, depth readout, and edge readout across the diffusion process. 
    A large portion of the image composition is already determined in the first 10\% of the diffusion process ($t=10$), as reflected by the crispness of our readouts which become further refined over time.
    }
    \label{fig:progressive_readouts}
    \vspace{-0.8cm}
\end{figure*}

In~\autoref{fig:progressive_readouts} we visualize the evolution of our readouts over the sampling process. The typical visualization of the denoising process, or the evolution of $x_t$ (row 1), can be relatively uninterpretable across a majority of the process. A more informative visualization is typically the intermediate prediction of the clean image (row 2). Our readouts (row 3-5) offer an additional tool for probing the diffusion process, derived from the features themselves rather than the sampling formulation. We find that the image composition is largely determined early in the diffusion process, where our readouts show a central figure in the foreground within the first step of sampling and a coarse silhouette within the first 10\%. Evidently, our readouts are useful for guidance because they are informative across the entire diffusion process, even at early noisy timesteps.

\subsection{Quantitative Comparison} 
In~\autoref{tab:pose_pck} we compare against fine-tuned conditional models on the task of pose control. We opt for using SDXL as our base model, since it significantly outperforms SDv1-5 in visual quality. Unfortunately, ControlNet~\cite{zhang2023adding} only has officially released weights on SDv1-5, so our SDXL comparisons are only with officially released checkpoints from T2IAdapter~\cite{mou2023t2i}. We do, however, provide additional supplementary comparisons to ControlNet on SDv1-5. %
In this evaluation, we use OpenPose~\cite{8765346} to extract the pose of the generated images from each method then compute the percentage of correct keypoints (PCK) against the input control. When a pose prediction has multiple instances, we take the highest scoring PCK across all combinations of instances between the input and synthesized image. We then average the PCKs across all samples. We compute the PCK at varying error thresholds, $\alpha \in \{0.05, 0.1, 0.2\}$. We also report the amount of added training data that was used to fine-tune the control method, as well as the added parameter count. Note that the parameter count of our method scales with the number of decoder layers (SDXL has 9 vs SDv1-5 has 12 layers), so our SDXL heads are actually more parameter-efficient than our SDv1-5 heads. SDXL trades off fewer decoder blocks with additional mid blocks, which we found in early experiments did not improve the quality of our spatially-aligned readouts. Finally, we see that our pose head trained on the vanilla base model can be transferred to guide a fine-tuned conditional model, further improving the performance of ControlNet~\cite{zhang2023adding} and T2IAdapter~\cite{mou2023t2i} by 13\% and 30\% PCK ($\alpha=0.05$) respectively. The compatibility between Readout Guidance and these fine-tuned conditional models derives from the fact that all methods keep the base model frozen, and therefore operate on the same common feature space as the original base model. 
\vfill\null
We also calculate image quality as measured by the CLIP Maximum Mean Discrepancy~\cite{jayasumana2023rethinking}. We compare the distributions of the real MSCOCO images, from which we derived the pose control and text prompt, with the generated images from each method. We opt to use CMMD as it is unbiased with respect to the evaluation sample size, unlike FID which is not meaningful on small image sets. As seen in~\autoref{tab:pose_cmmd}, our method actually slightly improves image quality when compared to the base model (No Control vs. RG$_{8.5k}$, T2IAdapter vs. T2IAdapter + RG$_{8.5k}$). We also see that training on more data leads to increasingly better image quality (RG$_{100}$ vs. RG$_{1k}$ vs. RG$_{8.5k}$). This likely explains why there is a leap in image quality when also using T2IAdapter, as it was trained on orders of magnitude more data ($\sim$3M images).

\subsection{Additional Examples} 
We show additional examples of spatially aligned control from our full method trained on all images in PascalVOC~\cite{pascal-voc-2012}. We show additional examples of pose control in~\autoref{fig:pose_suppl_a} and ~\autoref{fig:pose_suppl_b}, depth control in~\autoref{fig:depth_suppl_a} and ~\autoref{fig:depth_suppl_b}, and edge control in~\autoref{fig:edge_suppl_a} and ~\autoref{fig:edge_suppl_b}. In~\autoref{fig:pose_suppl_b}, we see that our pose readout can sometimes hallucinate extra detail, for example detecting fingers (col 2, row 4) or detecting the dog (col 1, row 2), even when they are not covered by the pose representation. Similarly, in~\autoref{fig:depth_suppl_a}, while our depth readout can reliably distinguish the foreground and background, it can struggle with producing a smooth and uniform depth across the entire object.
We show additional qualitative examples using our guidance on top of ControlNet~\cite{zhang2023adding} (\autoref{fig:controlnet_ours_suppl_a}) and T2I-Adapter~\cite{mou2023t2i} (\autoref{fig:t2iadapter_ours_suppl_a}). We also compare the variants of our model trained on 100, 1k, and 8.5k images in~\autoref{fig:train_size_suppl_a}. For simpler poses (col 1, row 8), all models perform equally well whereas for more difficult examples (col 1, row 4) our method trained on all 8.5k images performs the best.

\pagebreak
\section{Text Prompts}
\noindent We provide the text prompts used to generate the images in the main text below.
\begin{itemize}
\item \autoref{fig:dragdiff_synthetic}: \promptcaption{professional headshot of a woman with the eiffel tower in the background}, \promptcaption{photo of a squirrel with a hamburger at its feet}, \promptcaption{photo of a beagle hovering mid-air}, \promptcaption{a bear dancing on times square}
\item \autoref{fig:controlnet}: \promptcaption{two men on a tennis court shaking hands over the net
}, \promptcaption{a group of people playing with Frisbee's on the grass}, \promptcaption{a giraffe standing in a straw field next to shrubbery}, \promptcaption{two birds standing next to each other on a branch}, \promptcaption{a mountain goat stands on top of a rock on a hill}, \promptcaption{a man riding a horse followed by a dog}
\item \autoref{fig:image_var}: \promptcaption{A dog is walking down the street}, \promptcaption{An astronaut is skiing down the hill}
\item \autoref{fig:identity}: \promptcaption{oil painting half body portrait of a man on a city street in copenhagen}, \promptcaption{portrait of a man and his dog}, \promptcaption{photo of a female firefighter in the forest}, \promptcaption{photo of a woman at the zoo with a seal}
\item \autoref{fig:dragdiff_real}: \promptcaption{a photo of a tiger}, \promptcaption{a photo of a raccoon}, \promptcaption{a photo of a rabbit}, \promptcaption{a photo of a man holding a crocodile}, \promptcaption{an alpine ibex with horns}~\cite{alpineibex}, \promptcaption{wolf looking around}~\cite{wolflooking}
\item \autoref{fig:controlnet_ours}: \promptcaption{a woman that is sitting on a couch holding a remote}, \promptcaption{a young lady is playing a baseball bat game}, \promptcaption{a man wearing an apron peering into the bottom of an open fridge}, \promptcaption{a person on a surfboard on the water}
\item \autoref{fig:train_size}: \promptcaption{a person riding a scooter with folded cardboard}, \promptcaption{a group of people in a field playing frisbee}
\end{itemize}

\pagebreak
\clearpage

\anycolfigpdf{2}{figures/controlnet}{
  control_suppl.pdf/1/Pose Control,
  control_suppl.pdf/2/Pose Control
}{RG (Ours), Readout}{fig:pose_suppl_a}{\textbf{Pose Control}: Additional examples with our pose head.}{0.5cm}{-1cm}
\anycolfigpdf{2}{figures/controlnet}{
  control_suppl.pdf/3/Pose Control,
  control_suppl.pdf/4/Pose Control
}{RG (Ours), Readout}{fig:pose_suppl_b}{\textbf{Pose Control}: Additional examples with our pose head.}{0.5cm}{-1cm}

\anycolfigpdf{2}{figures/controlnet}{
  control_suppl.pdf/5/Depth Control,
  control_suppl.pdf/6/Depth Control
}{RG (Ours), Readout}{fig:depth_suppl_a}{\textbf{Depth Control}: Additional examples with our depth head.}{0.5cm}{-1cm}
\anycolfigpdf{2}{figures/controlnet}{
  control_suppl.pdf/7/Depth Control,
  control_suppl.pdf/8/Depth Control
}{RG (Ours), Readout}{fig:depth_suppl_b}{\textbf{Depth Control}: Additional examples with our depth head.}{0.5cm}{-1cm}

\anycolfigpdf{2}{figures/controlnet}{
  control_suppl.pdf/9/Edge Control,
  control_suppl.pdf/10/Edge Control
}{RG (Ours), Readout}{fig:edge_suppl_a}{\textbf{Edge Control}: Additional examples with our edge head.}{0.5cm}{-1cm}
\anycolfigpdf{2}{figures/controlnet}{
  control_suppl.pdf/11/Edge Control,
  control_suppl.pdf/12/Edge Control
}{RG (Ours), Readout}{fig:edge_suppl_b}{\textbf{Edge Control}: Additional examples with our edge head.}{0.5cm}{-1cm}

\anycolfigpdf{2}{figures/controlnet}{
  controlnet_ours_suppl.pdf/1/Pose Control,
  controlnet_ours_suppl.pdf/2/Pose Control
}{ControlNet, ControlNet + RG, Readout}{fig:controlnet_ours_suppl_a}{\textbf{Control Refinement}: Additional examples of \guidancename{} combined with  ControlNet~\cite{zhang2023adding}.}{0.1cm}{0cm}

\anycolfigpdf{2}{figures/t2iadapter}{
  t2iadapter_ours_suppl.pdf/1/Pose Control,
  t2iadapter_ours_suppl.pdf/2/Pose Control
}{T2IAdapter, T2IAdapter + RG, Readout}{fig:t2iadapter_ours_suppl_a}{\textbf{Control Refinement}: Additional examples of \guidancename{} combined with T2IAdapter~\cite{mou2023t2i}.}{0.1cm}{0cm}

\anycolfigpdf{2}{figures/controlnet}{
  train_size_suppl.pdf/1/Pose Control,
  train_size_suppl.pdf/2/Pose Control
}{100 Images, 1k Images, 8.5k Images}{fig:train_size_suppl_a}{\textbf{Limited Data}: Additional examples after training our pose head on varying numbers of images.}{0.2cm}{-0.5cm}

\end{document}